\documentclass[preprint,9pt]{elsarticle}

\usepackage[acronym]{glossaries}
\usepackage{longtable}
\usepackage{pdflscape}
\usepackage{float}
\usepackage{multirow}
\usepackage{amssymb}
\usepackage{xcolor}
\usepackage{tabularx}
\usepackage{makecell}
\usepackage{geometry}
\usepackage{graphicx}
\usepackage{subcaption}
\usepackage{amsmath}
\usepackage{geometry}
\usepackage{booktabs} % For better table lines
\usepackage{array} % For better column options
\usepackage{tikz}
\usepackage{pgfplots}
\pgfplotsset{compat=1.18}
\usepackage{xurl} % lets long URLs in the reference list break across lines
\usepackage[hidelinks]{hyperref}
\usepackage{adjustbox}
\usepackage{ragged2e}
\usepackage{amsthm}
\usepackage{framed} % Framing content
\usepackage{multicol} % Multiple columns environment
\usepackage{color,soul}
\usepackage{enumitem}
\usepackage{nomencl} % Nomenclature package
\makenomenclature
\begin{document}

\begin{frontmatter}

\title{Verifiable Logic in Medical LLMs: A Critical Review of Agentic Clinical Reasoning Optimization and Neuro-Symbolic Grounding}

\affiliation[inst1]{organization={Artificial Intelligence Department, University of Prince Mugrin},%Department and Organization
            %addressline={Address One}, 
            city={Medina },
            postcode={12435}, 
            %state={State One},
            country={Saudi Arabia}}
            
\affiliation[inst2]{organization={Robotics and Internet of Things Laboratory, Prince Sultan University},%Department and Organization
            %addressline={Address One}, 
            city={Riyadh },
            postcode={11586}, 
            %state={State One},
            country={Saudi Arabia}}
            
\author[inst1]{Youssef ElNahas}
\author[inst2]{Adel Ammar\corref{cor1}}
\ead{aammar@psu.edu.sa}
\author[inst1]{Soumaya Chaffar}
\author[inst2]{Wadii Boulila}

\cortext[cor1]{Corresponding author}

\begin{abstract}
Clinical large language models are usually judged by whether the final answer is right. Yet a model can reach a fluent, plausible output through invalid reasoning that static benchmarks do not detect, and confident hallucination, causally invalid inference, and spurious demographic shortcuts can pass conventional evaluation. This critical review covers four paradigms in turn: how clinical hallucination arises and why static metrics miss it; the agentic hallucination cascade, in which multi-agent and tool-augmented systems propagate and amplify early reasoning errors; process reward modeling as step-level supervision; and neuro-symbolic grounding, in which formal logic engines, ontology reasoners, and knowledge graphs place verifiable constraints on neural output. Within their encoded domains, systems with symbolic veto layers offer guarantees that hold relative to their encoded knowledge, and they report large gains on the safety measures they use. These results come from single studies, mostly in simulated or small-sample settings, and are not directly comparable with neural baselines. Process reward models remain bounded by the statistics of their training data. We propose a conceptual framework that maps each agentic failure mode to a symbolic intervention point, and we argue that clinical AI safety depends substantially on system architecture. Because current regulation emphasizes performance and does not require verification of reasoning steps, we argue that responsible certification should also audit reasoning trajectories alongside accuracy on static benchmarks.
\end{abstract}

%%Graphical abstract
%\begin{graphicalabstract}
%\includegraphics{grabs}
%\end{graphicalabstract}
% \begin{graphicalabstract}
% \centering
% %\includegraphics[scale=0.4]{xx.png}
% %\caption*{Visual overview of the paper structure depicting key sections and subsections.}
% \label{fig:graphical_diag}
% \end{graphicalabstract}

% Research highlights
% \begin{highlights}
% \item  xxxx.
% \item Txxx.
% \item xxx.
% \item xxx.
% \item xxx.
% \end{highlights}

\medskip
\begin{keyword}
%% keywords here, in the form: keyword \sep keyword
medical large language models \sep verifiable reasoning \sep clinical AI safety \sep process reward models \sep neuro-symbolic integration 

%% PACS codes here, in the form: \PACS code \sep code
%\PACS 0000 \sep 1111
%% MSC codes here, in the form: \MSC code \sep code
%% or \MSC[2008] code \sep code (2000 is the default)
%\MSC 0000 \sep 1111
\end{keyword}

\end{frontmatter}

%% \linenumbers

% Print glossary or acronyms list
%\printglossaries

%% main text

\section{Introduction}
\label{sec:intro}
Scaling and instruction tuning have pushed Large Language Models (LLMs) to passing scores on demanding medical exams such as the USMLE~\cite{Kung2023USMLE}, and they now summarize complex patient records fluently~\cite{van2024adapted}. They are also drifting from information retrieval toward clinical decision support, and there unsupported hallucination becomes a patient-safety problem. Clinical reasoning depends on structured thinking, both abductive and deductive, yet many LLMs are trained to reward the correct final answer regardless of how they reached it. This ``shortcut learning'' is hazardous in medicine. A missed allergy history or an overlooked contraindication can cause serious harm, so the reasoning itself has to be open to inspection and verification.

Recent work has followed two main directions. Agentic clinical workflows give LLMs autonomy, external tools such as medical calculators and electronic health records (EHRs), and multi-agent collaboration. Neuro-symbolic grounding instead constrains LLM output with formal knowledge structures, including Knowledge Graphs (KGs)~\cite{hogan2021knowledge} and medical ontologies such as UMLS~\cite{bodenreider2004unified}. The existing surveys (Table~\ref{tab:survey_comparison}) split along the same line. Lin et al.~\cite{lin2025survey}, Vatsal et al.~\cite{vatsal2026agentic}, and Wang et al.~\cite{wang2025survey} cover multi-agent architectures, planning strategies, and evaluation frameworks for clinical agents, yet they treat knowledge graphs and ontologies as retrieval resources rather than verification tools. Xu et al.~\cite{xu2025survey} and Gorenshtein et al.~\cite{gorenshtein2026neuro} discuss neuro-symbolic reasoning mostly inside single LLM-based systems and say little about dynamic multi-agent settings, where several agents interact. He et al.~\cite{he2025survey} acknowledge both directions but offer no framework that joins them.

\begin{table}[H]
\centering
\caption{\textbf{Comparative Analysis of Existing Surveys on Medical LLMs.}}
\label{tab:survey_comparison}
\footnotesize
\setlength{\tabcolsep}{4.5pt}
\renewcommand{\arraystretch}{1.05}

\begin{tabularx}{\textwidth}{
    >{\raggedright\arraybackslash}p{1.9cm}
    >{\raggedright\arraybackslash}p{1.6cm}
    >{\raggedright\arraybackslash}p{1.8cm}
    >{\centering\arraybackslash}p{1.6cm}
    >{\centering\arraybackslash}p{1.6cm}
    >{\raggedright\arraybackslash}X
}
\toprule
\textsc{Study} & 
\textsc{Method} & 
\textsc{Scope} & 
\textsc{Agentic} & 
\textsc{Neuro-Sym.} & 
\textsc{Limitations} \\
\midrule

% Row 1: Lin et al.
Lin et al. (2025) \cite{lin2025survey} &
Survey (structured search) \& Taxonomy &
Taxonomy of medical LLM multi-agent systems. &
Core Focus. &
Partial. &
Does not provide a unified framework in which symbolic reasoning constrains agent outputs. \\
\cmidrule{1-6}

% Row 2: Vatsal et al.
Vatsal et al. (2026) \cite{vatsal2026agentic} &
Survey \& Taxonomy &
Evaluation taxonomy for medical LLM agents. &
Core Focus. &
Partial. &
Does not incorporate neuro-symbolic integration or mechanisms to audit dynamic reasoning. \\
\cmidrule{1-6}

% Row 3: Wang et al.
Wang et al. (2025) \cite{wang2025survey} &
Survey &
Architecture and challenges of clinical LLM agents. &
Core Focus. &
Partial. &
Treats retrieval and symbolic grounding as equivalent and does not include a formal constraint layer. \\
\cmidrule{1-6}

% Row 4: Xu et al.
Xu et al. (2025) \cite{xu2025survey} &
Survey \& Framework &
Unification of LLMs and Knowledge Graphs. &
Partial. &
Core Focus. &
Does not extend knowledge graph verification to multi-agent systems. \\
\cmidrule{1-6}

% Row 5: Gorenshtein et al.
Gorenshtein et al. (2026) \cite{gorenshtein2026neuro} &
Systematic Review (PRISMA 2020) \& Taxonomy &
Taxonomy of neuro-symbolic integration patterns. &
Partial. &
Core Focus. &
Views symbolic components primarily as static verifiers rather than active participants in agent workflows. \\
\cmidrule{1-6}

% Row 6: He et al.
He et al. (2025) \cite{he2025survey} &
Survey &
Roadmap of generative LLMs in healthcare. &
Mentioned. &
Partial. &
Does not examine how dynamic agents can be integrated with rigorous neuro-symbolic verification. \\
\cmidrule{1-6}

% Row 7: This Review
\textbf{This Review} &
\textbf{Critical Review} &
\textbf{Integrated Framework} &
\textbf{Core Focus} &
\textbf{Core Focus} &
\textbf{Purposive selection; the proposed integrated framework has not yet been validated in clinical settings.} \\
\bottomrule

\end{tabularx}
\end{table}

This split leaves a practical question open: how can the strengths of both approaches be combined for safe clinical deployment? Agentic systems, even with sophisticated multi-agent collaboration, remain exposed to ``hallucination cascades,'' in which agents reinforce one another's reasoning errors because nothing external and verifiable checks them. Static neuro-symbolic pipelines have the opposite profile. Their consistency and auditability are strong, but latency and outdated knowledge limit them, and unstructured clinical narratives complicate their application. The two literatures have mostly ignored each other. To our knowledge, no earlier survey asks how autonomous multi-agent systems can work with formal symbolic reasoning so that verification constrains every stage of clinical reasoning, from forming hypotheses to recommending treatment, while the whole process stays auditable against established medical ontologies.

We address that divide with a unified framework, Verifiable Logic in Medical LLMs. Our position has two parts. Autonomous clinical agents cannot safely rely on neural self-reflection alone, and formal medical ontologies cannot support modern clinical workflows if they stay confined to static, pipeline-style reasoning. We argue that neuro-symbolic grounding should be the verifiable foundation that guides and constrains agentic reasoning across the whole decision-making process.

The main contributions of this critical review are as follows:
\begin{enumerate}
\item \textbf{The hallucination cascade.} We characterize the cascade as a failure mode distinct from single-model hallucination. Earlier work showed that a single model's early mistakes can snowball into later ones~\cite{Zhang2024Snowball}, and concurrent work has begun to measure error propagation across multi-agent LLM pipelines~\cite{Jamshidi2026Cascade}. We bring these observations together for clinical workflows, where errors spread across agents and tools, accumulate, and become harder to detect, and where outcome-based evaluation does not capture them.

\item \textbf{The agentic-symbolic divide.} We argue that the split between the two literatures is itself a barrier to safe deployment. To our knowledge, no earlier survey joins them in one verifiable framework.

\item \textbf{The architecture argument.} Across four paradigms, we argue that clinical AI safety depends substantially on system architecture, and not only on training, calibration, or prompting. Within their formal domains, symbolic veto mechanisms offer guarantees that the reviewed neural approaches do not, although those guarantees hold only relative to the encoded knowledge and the facts supplied to them, and we argue that formal constraint enforcement should be part of safe clinical AI.
\end{enumerate}

These contributions have a regulatory implication, developed in Section~\ref{sec:conclusion}: current frameworks emphasize performance and post-market monitoring and do not require verification of the reasoning process, which we argue reliable certification of clinical AI also needs.

Four research questions guide the synthesis in Sections~\ref{sec:core_problem}--\ref{sec:neurosymbolic}:

\begin{enumerate}[label=\textbf{RQ\arabic*.}]
\item What types of clinical LLM hallucinations exist, and why do current outcome-based evaluation frameworks fail to detect them?
\item How do agentic clinical architectures create and amplify hallucinations through cascade failures, which differ in kind from the errors of a single model?
\item How far can Process Reward Models verify clinical reasoning, and where does that verification stop?
\item What safety benefits do neuro-symbolic grounding architectures offer over purely neural approaches, and under what conditions do those benefits hold?
\end{enumerate}

The word \textit{verifiable} is used loosely in the literature, so we fix five verification regimes and apply them consistently. \textit{Deductive verification} means a claim is formally entailed by a set of premises under a specified inference system, as in Prolog backward chaining. The guarantee is conditional: a derivation is sound only relative to its axioms and to the facts supplied as premises, so it is only as good as the completeness and correctness of both, and of the procedure that encoded them \cite{Tan2026ProCDS,Quan2024TheoremProving}. \textit{Ontology-based verification} uses a description logic reasoner such as HermiT or Pellet over a formal biomedical ontology (OWL/RDF, possibly with SWRL rules) to perform one of two different tasks. A consistency check accepts a claim when adding it to the ontology produces no contradiction; an entailment check accepts it only when the ontology implies it. Consistency is the weaker test, since a claim can be compatible with everything the ontology records without being supported by it, so ontology-based verification is a special case of deductive verification only when entailment is tested. In either case, computational cost rises with the ontology's expressiveness \cite{OWLHermiT2025}. \textit{Evidence-grounded verification} supports claims with material retrieved from external sources such as clinical guidelines, knowledge bases, or structured EHRs. It gives no formal guarantee of entailment, but the reasoning stays transparent and auditable, and the evidence can be updated without retraining \cite{Yun2025MedPRM,Sohn2026PRA,OntologyCOVID2026}. \textit{Probabilistic learned verification} uses a trained discriminative model to score reasoning steps or outputs. The guarantees are statistical, not formal, and they depend heavily on the quality and coverage of the annotated training data \cite{Wang2024ClinicalNotePRM,Thomas2025RadiologyPRM}. \textit{Structural process-level verification} is a by-product of training objectives that encourage structured reasoning, chain-of-thought supervision being the usual example. It pushes models toward process-oriented behavior without amounting to formal verification \cite{Wu2025MedReason,Mu2025MedCEG,Rui2025MedCCO}.

Throughout this review we keep four properties apart, because the literature often runs them together. \textit{Factual provenance} asks whether an extracted fact is actually present in the patient record. \textit{Consistency} asks whether a claim contradicts the encoded knowledge. \textit{Entailment} asks whether the encoded knowledge, together with the extracted facts, implies the claim. \textit{Clinical validity} asks whether the claim is correct for the patient. A formal verifier can establish consistency or entailment, but only relative to its premises, its encoding, and its inference procedure. It says nothing about provenance unless facts are checked against their source, and nothing about clinical validity unless the encoded knowledge is itself correct and current.

The paper is organized as follows. Section~\ref{sec:methods} describes the review methodology. Section~\ref{sec:core_problem} presents a taxonomy of clinical hallucinations and shows why outcome-based metrics miss failures in clinical reasoning. Section~\ref{sec:agentic} turns to multi-agent clinical workflows and the risk of hallucination cascades. Section~\ref{sec:prm} evaluates Process Reward Modeling as a candidate fix and where it falls short in medical applications. Section~\ref{sec:neurosymbolic} reviews neuro-symbolic grounding and develops the proposed architectural synthesis. Section~\ref{sec:conclusion} closes with the main findings, the open research problems, and the limits of this review.

\section{Review Methodology}
\label{sec:methods}

\subsection*{Design and Scope}

This article is a critical narrative review that maps the architectural approaches to clinical LLM safety across the four paradigms defined by RQ1--RQ4. To keep the process transparent, the search and selection steps are reported following the structure of the PRISMA Extension for Scoping Reviews (PRISMA-ScR)~\cite{Tricco2018PRISMAScR}. Because the review was conducted without a pre-registered protocol, it is presented as a narrative synthesis and not as a formal scoping review. The studies were purposively selected on the basis of how clearly they describe their architecture, explain how the system works, and relate to these paradigms. This design favors depth of mechanistic analysis over exhaustive coverage, so the retained studies should be read as a representative evidence base for each paradigm.

\subsection*{Search Strategy}

Literature was identified with structured prompt-based queries sent to five AI-assisted retrieval tools: Perplexity AI (live indexed web retrieval), Claude (Anthropic), DeepSeek, Grok (xAI), and Kimi, with one query per research question. Queries for RQ1 and RQ2 were submitted on 8 April 2026, and queries for RQ3 and RQ4 on 28 April 2026. The search terms for each are listed below and were submitted to each tool as written.

{\raggedright
\noindent\textbf{RQ1 -- Hallucination taxonomy and benchmark failure:}\\
\texttt{("Large Language Models" OR "LLM") AND ("healthcare" OR "medicine") AND ("clinical hallucinations" OR "shortcut learning" OR "context omission" OR "RLHF miscalibration" OR "static metrics" OR "outcome-based evaluation")} -- 2024 to 2026\par}

\smallskip
{\raggedright
\noindent\textbf{RQ2 -- Agentic architectures and cascade failure modes:}\\
\texttt{("LLM" OR "large language model") AND ("autonomous agents" OR "multi-agent systems") AND ("clinical" OR "medical") AND ("compounding errors" OR "hallucination cascade" OR "ReAct" OR "multi-agent debate" OR "EHR" OR "RAG")} -- 2024 to 2026\par}

\smallskip
{\raggedright
\noindent\textbf{RQ3 -- Process reward model verification:}\\
\texttt{("Process Reward Model" OR "PRM" OR "process supervision" OR "step-by-step reasoning") AND ("medical" OR "healthcare" OR "clinical") AND ("reasoning verification" OR "intermediate steps")} -- 2024 to 2026\par}

\smallskip
{\raggedright
\noindent\textbf{RQ4 -- Neuro-symbolic grounding:}\\
\texttt{("neuro-symbolic" OR "knowledge graph" OR "symbolic logic" OR "rule engine" OR "Prolog") AND ("LLM" OR "large language model") AND ("clinical" OR "medicine") AND ("verification" OR "hallucination" OR "veto")} -- 2024 to 2026\par}

\smallskip
\noindent All retrieved records were screened by hand. Titles, abstracts, and, where needed, full texts were checked to confirm that each article existed, that its content was accessible, and that it is relevant. Records that could not be verified in indexed sources such as IEEE Xplore, the ACM Digital Library, PubMed, or Google Scholar were excluded, however they were found. Verified records were logged in a charting spreadsheet with one sheet per research question, giving 140 records, or 138 after two records that appeared under two research questions were merged. Figure~\ref{fig:prisma} summarizes the selection, and Supplementary Table~S1 lists the included studies.

% Set the included count once; the full-text exclusions are computed from it.
\newcommand{\nLogged}{140}
\newcommand{\nDuplicates}{2}
\newcommand{\nAssessed}{\the\numexpr\nLogged-\nDuplicates\relax}
\newcommand{\nIncluded}{98}   % must match the total in Table 2
\newcommand{\nExcluded}{\the\numexpr\nAssessed-\nIncluded\relax}

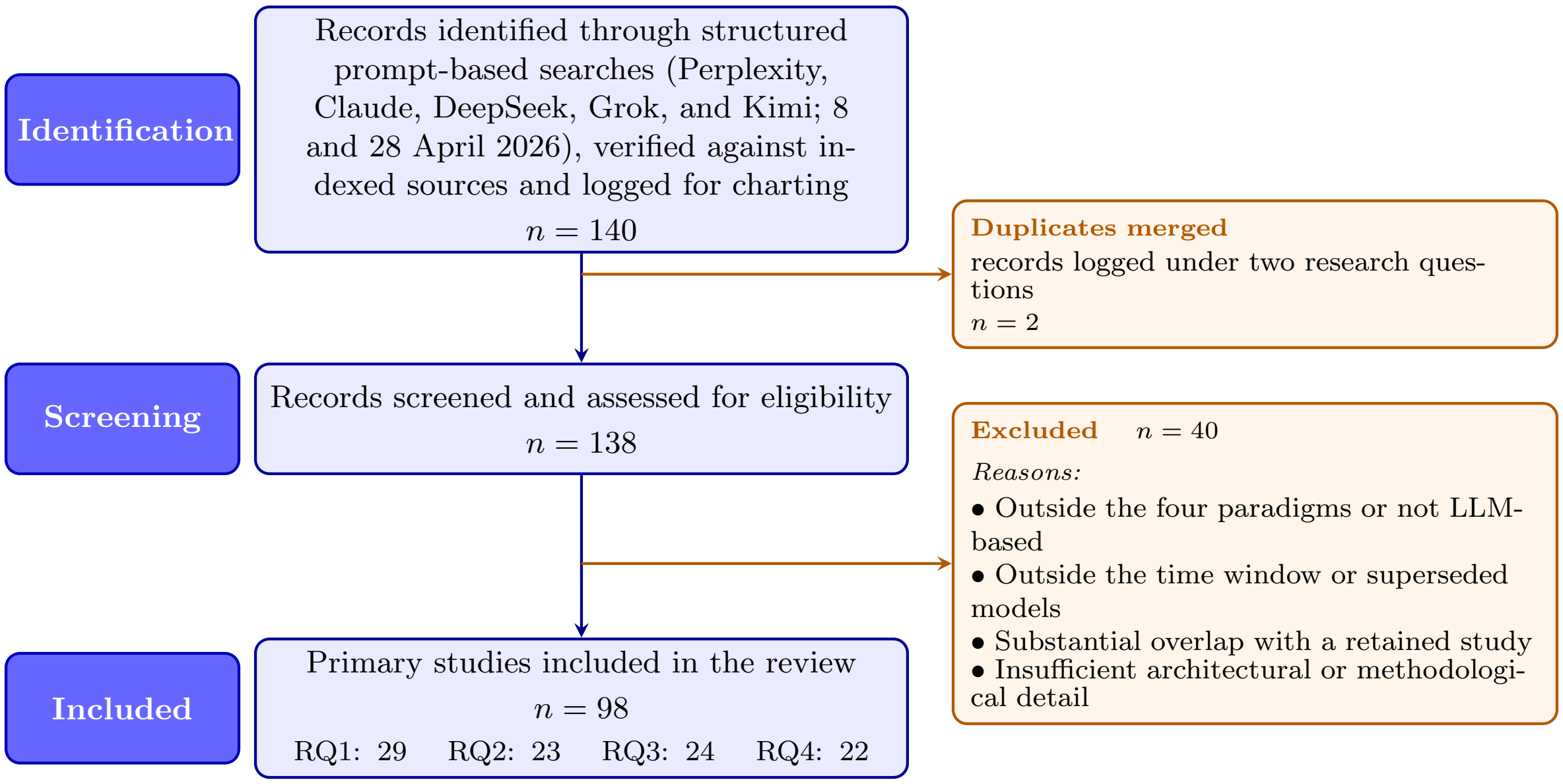
\begin{figure}[H]
\centering
\begin{adjustbox}{max width=\textwidth, max height=0.5\textheight}
\begin{tikzpicture}[
  mbox/.style={rectangle, draw=blue!60!black, line width=0.8pt, rounded corners=4pt,
               minimum width=6.5cm, text width=6.3cm,
               minimum height=1.1cm, fill=blue!8,
               font=\small, align=center},
  ebox/.style={rectangle, draw=orange!70!black, line width=0.8pt, rounded corners=4pt,
               minimum width=5.9cm, text width=5.7cm,
               fill=orange!8, font=\scriptsize, align=left,
               inner sep=5pt},
  pbox/.style={rectangle, draw=blue!70!black, line width=0.8pt, rounded corners=4pt,
               minimum width=2.3cm, text width=2.1cm,
               minimum height=1.1cm, fill=blue!60,
               font=\small\bfseries, align=center, text=white},
  arr/.style={-stealth, thick, blue!50!black},
  arrx/.style={-stealth, thick, orange!70!black},
]
\node[pbox] at (-4.6,  0.0) {Identification};
\node[pbox] at (-4.6, -2.9) {Screening};
\node[pbox] at (-4.6, -5.8) {Included};

\node[mbox] (n1) at (0,  0.0) {
  Records identified through structured prompt-based searches
  (Perplexity, Claude, DeepSeek, Grok, and Kimi; 8 and 28 April 2026),
  verified against indexed sources and logged for charting\\[2pt]
  \textbf{\mbox{$n=\nLogged$}}
};
\node[mbox] (n2) at (0, -2.9) {
  Records screened and assessed for eligibility\\[2pt]
  \textbf{\mbox{$n=\nAssessed$}}
};
\node[mbox] (n3) at (0, -5.8) {
  Primary studies included in the review\\[2pt]
  \textbf{\mbox{$n=\nIncluded$}}\\[1pt]
  {\footnotesize RQ1: 29 \quad RQ2: 23 \quad RQ3: 24 \quad RQ4: 22}
};

\node[ebox] (e1) at (6.75, -1.45) {
  \textcolor{orange!70!black}{\textbf{Duplicates merged}}\\[2pt]
  {\footnotesize records logged under two research questions}\\[1pt]
  \textbf{$n = \nDuplicates$}
};
\node[ebox] (e2) at (6.75, -4.35) {
  \textcolor{orange!70!black}{\textbf{Excluded}} \quad \textbf{$n = \nExcluded$}\\[4pt]
  \textit{Reasons:}\\[1pt]
  {\footnotesize
  $\bullet$ Outside the four paradigms or not LLM-based\\
  $\bullet$ Outside the time window or superseded models\\
  $\bullet$ Substantial overlap with a retained study\\
  $\bullet$ Insufficient architectural or methodological detail}
};

\draw[arr] (n1) -- (n2);
\draw[arr] (n2) -- (n3);
\draw[arrx] (0,-1.45) -- (e1.west);
\draw[arrx] (0,-4.35) -- (e2.west);
\end{tikzpicture}
\end{adjustbox}
\caption{Study selection flow, presented in the format of the PRISMA-ScR flow diagram~\cite{Tricco2018PRISMAScR} for transparency. Records were retrieved through structured prompt-based searches with five AI-assisted retrieval tools and verified by hand against indexed sources; counts start from the verified records logged for charting. The included studies are all primary studies and are listed in Supplementary Table~S1; prior surveys, precursor works and regulatory or background sources are cited separately. Exclusion reasons correspond to the eligibility criteria and temporal scope described in this section.}
\label{fig:prisma}
\end{figure}

\subsection*{Temporal Scope}

The review uses a two-tier time frame. Secondary literature (earlier survey papers analyzed in the gap analysis of Section~\ref{sec:intro} and Table~\ref{tab:survey_comparison}) was drawn from January 2023 to April 2026, to show how the field has developed over time. Primary literature (specific system implementations, benchmarks, multi-agent architectures, process reward models, and neuro-symbolic frameworks evaluated in Sections~\ref{sec:core_problem}--\ref{sec:neurosymbolic}) was limited to April 2024 to April 2026. There is one exception, for foundational methodological precursors that directly support in-scope architectures. Papers published before April 2024 are included where they provide the algorithmic basis of a reviewed system. Examples are the ReAct framework~\cite{Yao2023ReAct} and the multi-agent debate paradigm~\cite{Du2023MultiAgentDebate} (both 2023), which underpin the agentic architectures of Section~\ref{sec:agentic}, and the DeepProbLog, Scallop, and Logic Tensor Network architectures cited in Section~\ref{sec:neurosymbolic}. These precursor papers are identified on first use.

\subsection*{Inclusion Criteria}
\label{subsec:eligibility}

Papers were included only if they met all of the following conditions: (1) they addressed at least one of the four review paradigms (clinical hallucination taxonomy, agentic workflow architecture, process reward modeling, or neuro-symbolic grounding); (2) they reported empirical evaluation results, a formal analysis, or an architectural description detailed enough for mechanistic assessment; (3) they appeared in a peer-reviewed venue or, if they were preprints, gave enough methodological detail for their claims to be evaluated independently; and (4) for primary architectural studies, they fell within the time window above or qualified as a foundational precursor.

\subsection*{Exclusion Criteria}
Papers were excluded if they: (1) reported only overall benchmark scores without describing the underlying reasoning architecture, a failure mode analysis, or a mechanistic account of the performance; (2) evaluated clinical LLM performance only through outcome-based accuracy metrics and did not address process-level or architectural factors; (3) overlapped substantially with a system already represented by a retained paper; or (4) did not give enough methodological detail for their claims to be verified independently. Of the 138 records assessed, 40 were excluded under these criteria or because they fell outside the paradigms or time window defined above.

\subsection*{Data Charting \& Source Classification}
Each logged record was charted with its title, authors, venue, year and identifier, together with fields specific to its research question: for RQ1, the type of error or hallucination, a clinical example, and the computational explanation of the failure; for RQ2, the agent architecture, the tools available to the agent, and where the system failed or errors propagated; for RQ3, the clinical task, the reinforcement learning or reward architecture, and the evaluator of reasoning steps; and for RQ4, the symbolic engine or knowledge source, the verification strategy, and the trade-off between safety gain and cost. During synthesis, each included study was also assigned a verification regime (following the five regimes defined in Section~\ref{sec:intro}) and a source type, and its empirical results and reported limitations were taken from the study itself. Papers that contribute to more than one paradigm were charted under each but are counted once in Table~\ref{tab:paradigm_characteristics} and Supplementary Table~S1. This charted information underpins the taxonomy tables at the end of each main section. Retained papers are sorted into three source types: \textit{peer-reviewed published works} (appearing in indexed journals or conference proceedings at the time of retrieval); \textit{preprints} (posted on arXiv, medRxiv, bioRxiv, or similar, without confirmed peer-review acceptance at the time of retrieval, and included where the methodological detail allows independent assessment); and \textit{conceptual or speculative frameworks} (proposing architectural designs or theoretical arguments without empirical validation of the complete system). These are labeled as such where cited, and they are not used as the main evidence for quantitative safety claims. Source types reflect the version available at retrieval. Where a screened preprint was published after the search closed, the reference list gives the published version for completeness, the source-type label still reflects its status at retrieval, and Supplementary Table~S1 identifies the version screened. Systems evaluated only outside the clinical domain are marked with \textsuperscript{\dag} in Tables~\ref{tab:paradigm_characteristics} and~\ref{tab:verification_regimes}; they are included for their verification mechanism, not as clinical evidence.

\subsection*{Selection Bias Mitigation}

Because studies were selected purposively, three measures were taken to limit selection bias in the synthesis. First, the inclusion and exclusion criteria are stated explicitly (Section~\ref{subsec:eligibility}) and were applied to every retrieved record in the same way, whether or not a study supported the architectural thesis of this review. Second, the review deliberately sought out and kept contrary evidence, including studies that document failure modes of neuro-symbolic approaches (Sections~\ref{subsubsec:ns_tradeoffs} and~\ref{subsec:ns_eval}) and limitations of process reward models (Section~\ref{subsec:eval_bottleneck}). Third, Table~\ref{tab:survey_comparison} sets this review against earlier surveys that use different methods and scopes, including surveys whose conclusions do not fully agree with the unified framework proposed here. These measures reduce but do not remove the risk of selection bias. Its remaining effect on the conclusions is discussed in Section~\ref{sec:conclusion}.

\begin{table}[H]
\footnotesize
\setlength{\tabcolsep}{5pt}
\renewcommand{\arraystretch}{1.2}
\caption{Characteristics of the 98 included primary studies, grouped by research paradigm (RQ1--RQ4). Each study is counted once, under the paradigm that matches its main contribution; studies that also bear on other paradigms are cited there in the text but are not counted again. Prior surveys, precursor works and regulatory or background sources are excluded from the counts. The full list of included studies is given in Supplementary Table~S1. Superscripts show the source type: P = peer-reviewed, PP = preprint, CF = conceptual or speculative framework; \dag\ = evaluated outside the clinical domain.}
\label{tab:paradigm_characteristics}
\begin{tabularx}{\linewidth}{p{0.13\linewidth} p{0.04\linewidth} p{0.18\linewidth} p{0.16\linewidth} p{0.19\linewidth} X}
\toprule
\textbf{Paradigm} &
\textbf{n} &
\textbf{Dominant Regime(s)} &
\textbf{Dominant Metric Type} &
\textbf{Representative Systems} &
\textbf{Key Unresolved Gap} \\
\midrule

RQ1: Hallucination taxonomy &
29 &
Not applicable (diagnostic, not verification-layer) &
Hallucination or omission rate, F1 &
mARC-QA\textsuperscript{P}~\cite{mARC2025}, ShatterMed-QA\textsuperscript{PP}~\cite{Zi2026}, MedHallu\textsuperscript{PP}~\cite{MedHalluPandit2025}, MedCaseReasoning\textsuperscript{PP}~\cite{WuMedCase2025} &
Few benchmarks test counterfactual demographic invariance or adversarial robustness \\

\midrule

RQ2: Agentic cascade &
23 &
Not applicable (architectural, not verification-layer) &
Task success rate, cascade or error-propagation rate &
EHRAgent\textsuperscript{P}~\cite{Shi2024EHRAgent}, MedAgentBench\textsuperscript{P}~\cite{Jiang2025MedAgentBench}, MedAgentAudit\textsuperscript{PP}~\cite{MedAgentAudit2025}, TrustEHRAgent\textsuperscript{PP}~\cite{Song2025TrustEHRAgent} &
Calibration cliff: very low correct-answer yield under stringent reliability requirements \\

\midrule

RQ3: PRM verification &
24 &
Probabilistic learned, Structural process-level &
Accuracy gain, F1, reasoning recall &
Med-PRM\textsuperscript{P}~\cite{Yun2025MedPRM}, VPRM\textsuperscript{PP}~\cite{Pronesti2026VPRM}, MedCEG\textsuperscript{PP}~\cite{Mu2025MedCEG}, MedS$^3$\textsuperscript{P}~\cite{Jiang2025MedS3} &
Evaluation infrastructure lags system development; calibration not integrated into PRM benchmarks \\

\midrule

RQ4: Neuro-symbolic grounding &
22 &
Deductive, Ontology-based, Evidence-grounded &
Hallucination reduction (\%), protocol adherence (\%) &
ProCDS\textsuperscript{P}~\cite{Tan2026ProCDS}, Ontology-grounded GraphRAG\textsuperscript{P}~\cite{OntologyCOVID2026}, GraphEval\textsuperscript{P\,\dag}~\cite{Sansford2024GraphEval}, Explanation-Refiner\textsuperscript{P\,\dag}~\cite{Quan2024TheoremProving} &
Specification soundness, veto precision/recall, and temporal specification decay are unaudited \\

\bottomrule
\end{tabularx}
\renewcommand{\arraystretch}{1}
\end{table}

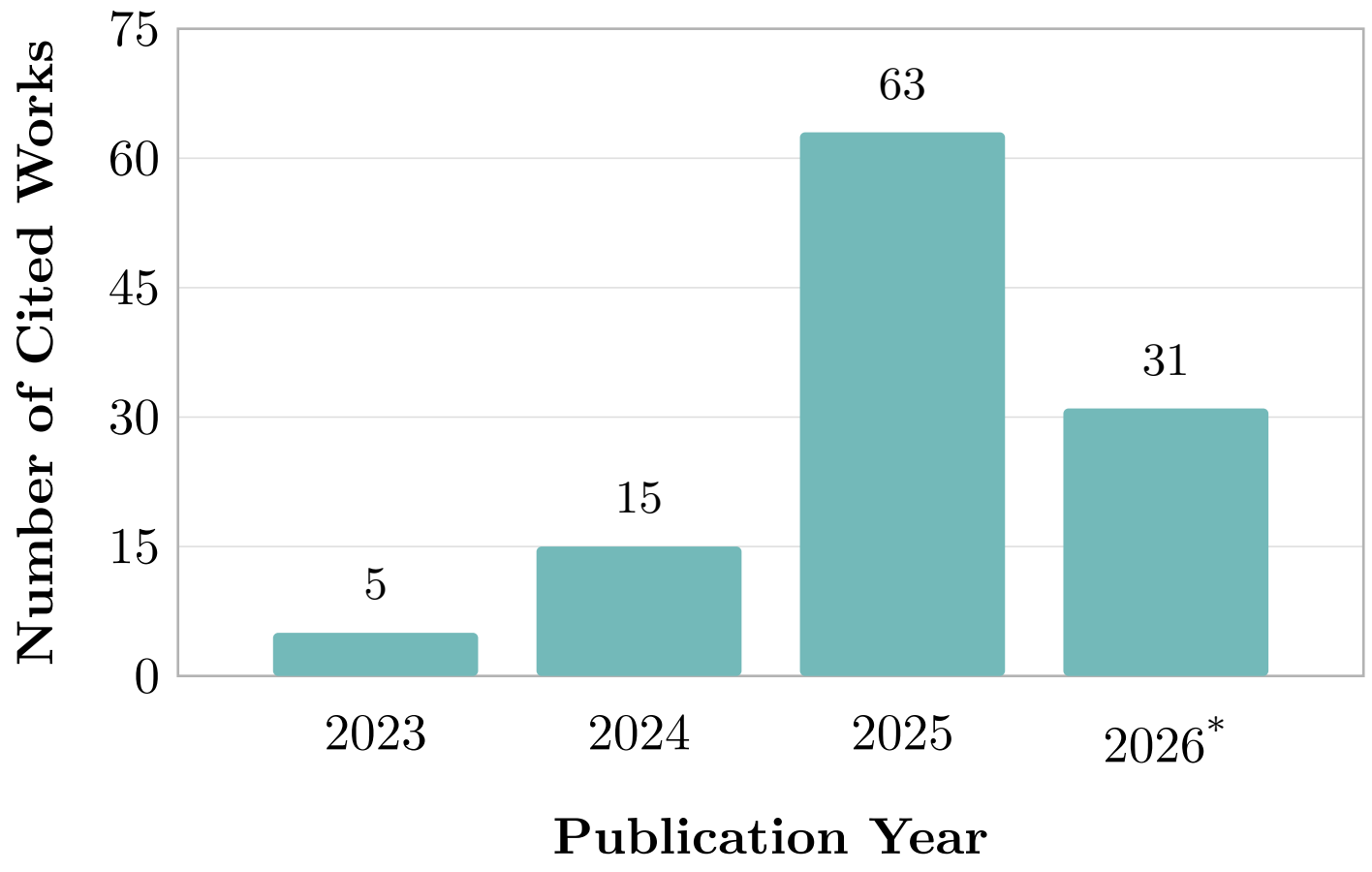
\begin{figure}[H]
\centering
\begin{tikzpicture}
\begin{axis}[
    ybar,
    bar width=1.4cm,
    width=0.55\textwidth,
    height=6cm,
    xlabel={\textbf{Publication Year}},
    ylabel={\textbf{Number of Cited Works}},
    xtick={1,2,3,4},
    xticklabels={2023, 2024, 2025, {2026\textsuperscript{*}}},
    ymin=0, ymax=75,
    ytick={0,15,30,45,60,75},
    ymajorgrids=true,
    grid style={line width=0.3pt, draw=gray!25},
    axis line style={line width=0.5pt, draw=gray!60},
    tick style={draw=none},
    enlarge x limits=0.25,
    nodes near coords,
    nodes near coords align={vertical},
    every node near coord/.append style={
        font=\small\bfseries,
        color=black,
        yshift=3pt
    },
    axis background/.style={fill=white},
    xlabel style={yshift=-4pt},
    ylabel style={yshift=4pt},
]
\addplot[
    fill=teal!55,
    draw=none,
    rounded corners=1pt
] coordinates {
    (1, 5)
    (2, 15)
    (3, 63)
    (4, 31)
};
\end{axis}
\end{tikzpicture}
\caption{Publication years of the works cited in this review, 2023--2026 ($N = 114$ of 129 references). The sharp rise in 2025 reflects how recent the literature on clinical LLM safety and verification is. Not shown are 15 earlier works (foundational methods, description-logic complexity results, reporting guidelines, earlier regulatory texts including HIPAA and the GDPR, and the DSM-5-TR). Each work is counted by the year given in its reference entry, which for preprints that have since been published is the year of publication. Unlike Table~\ref{tab:paradigm_characteristics}, the figure covers all cited works, not only the 98 included primary studies. \textsuperscript{*}The 2026 count is partial because the search closed on 28 April 2026; one later preprint is cited as concurrent work.}
\label{fig:papers_by_year}
\end{figure}

\section{Clinical Hallucinations and the Limits of Outcome-Based Metrics}
\label{sec:core_problem}

A model can score very high on standardized medical benchmarks and still make errors no clinician would accept, so a high score says little on its own about whether it reasons safely. This section sorts clinical hallucinations into four related categories (Section~\ref{subsec:anatomy}) and then asks why outcome-based evaluation can miss some of them (Section~\ref{subsec:static_failure}).

\subsection{Clinical Reasoning Dynamics: The Anatomy of Hallucinations}
\label{subsec:anatomy}

\textit{Hallucination} here means model output that sounds convincing but has no support in the available evidence. In medicine the cost is more than a wrong sentence. A hallucinated drug dose, an invented laboratory value, or a missed contraindication can cause serious patient harm. To see \textit{why} LLMs hallucinate in clinical settings, we separate the problem into at least four subtypes, each with its own mechanism.

\subsubsection*{Shortcut Learning and Spurious Correlations}
\label{subsubsec:shortcut}

The first failure mode starts with flawed knowledge representation. Models lean on statistical shortcuts learned from common patterns, and they break on uncommon ones. The mARC-QA benchmark~\cite{mARC2025} exposes this shallow pattern matching by borrowing the \textit{Einstellung} cognitive bias~\cite{Luchins1942} to probe inflexible reasoning on ``long-tail'' vignettes. ShatterMed-QA~\cite{Zi2026} shows that models exploit generic ``hub nodes'' in knowledge graphs to skip genuine multi-hop clinical reasoning. When these shortcuts are removed, models do not fail at random: up to 53\% of a model's errors fall on a topology-driven distractor, against a 33\% random baseline. When the masked intermediate entity is supplied through retrieval, models recover up to nearly 70\% of their zero-shot failures, although recovery varies widely and falls to 7\% for one medical model. The authors therefore attribute most failures to gaps in the model's knowledge more than to a pure reasoning deficit~\cite{Zi2026}.

Demographic stereotypes get encoded as false clinical correlations too. With clinical details held constant, unhoused patients evaluated by GPT-4o were 73.3\% more likely than controls to be recommended a mental health assessment ($p<0.001$)~\cite{OmarNatMed2025}. Standard benchmarks miss effects like this (Section~\ref{subsec:static_failure}) because few of them test counterfactual demographic invariance, that is, whether the answer stays the same when only the patient's demographics change. Kim et al.~\cite{Kim2025hallucinations} place such errors within a five-cluster hallucination taxonomy, one cluster of which is spurious correlations. Their physician audits attribute 64 to 72\% of residual hallucinations to failures of causal or temporal reasoning rather than to knowledge gaps. Medical specialization does not remove these errors: in the same analysis, general-purpose models outperformed medical-specialized ones (76.6\% vs.\ 51.3\% median hallucination-free rate)~\cite{Kim2025hallucinations}.

\subsubsection*{Context Omission and Adversarial Fragility}
\label{subsubsec:omission}

Context omission, the second failure mode, gets overlooked even though dropping a contraindication or a red-flag symptom can cause serious harm. In consultation note generation, Asgari et al.~\cite{Asgari2025} found hallucinations in 1.47\% of 12,999 generated note sentences and omissions of 3.45\% of 49,590 transcript sentences; omissions were more frequent, but 44\% of the hallucinated sentences were clinically major, compared with 16.7\% of the omitted ones. Measured per document instead of per sentence, errors are common: of GPT-4 summaries of 100 Emergency Department visits~\cite{Williams2025}, 47\% omitted clinically relevant information and 42\% contained at least one hallucination, although individual errors were judged to carry low potential for harm on average. Normand et al.~\cite{Normand2025} describe a related ``protocol vs patient gap'' in medication safety reviews on real NHS primary-care records: the system applied standard guidelines without adjusting for individual context such as frailty or goals of care, and the authors attribute most failures to contextual reasoning rather than missing medication knowledge.

Adversarial and conversational inputs make this weakness worse, because models cannot tell clinical ground truth from whatever the user asserts. When Omar et al.~\cite{OmarCommMed2025} planted single fabricated details in clinical vignettes, hallucination rates rose to between 50\% and 82\%. Prompt-based mitigation cut the mean rate only from 66\% to 44\%, and changing the temperature did nothing. The models elaborate on the planted misinformation, a form of conversational sycophancy. Li et al.~\cite{Li2026} show that models consistently validate medically unsound patient self-diagnoses and fail to resolve contradictory statements.

\subsubsection*{Miscalibration and Overconfidence}
\label{subsubsec:rlhf}

Preference-based fine-tuning such as RLHF is widely thought to discourage hedging, because confident answers tend to be preferred by raters; whatever the cause, current models often guess instead of admitting ignorance, a dangerous trade in clinical settings. Griot et al.~\cite{Griot2025} observe this directly: on unanswerable queries, models give a confident best guess instead of flagging uncertainty. Omar et al.~\cite{OmarJMIR2025} show how weak the link between confidence and correctness can be: even GPT-4o, the model whose confidence best separated correct from incorrect answers, showed a gap of only 5.4\% between the two, so stated confidence carries little information about correctness. Savage et al.~\cite{Savage2025} traced the problem specifically to verbal self-report. Sample consistency does give accurate calibration, but it needs multiple forward passes per query, a cost most clinical deployments cannot absorb.

Conversation makes the same miscalibration more dangerous than factual recall does. Au Yeung et al.~\cite{Yeung2025} tested models in simulated psychosis scenarios, where they not only failed to catch delusional content but actively confirmed it (mean score $0.91 \pm 0.88$) and enabled harmful reasoning ($0.69 \pm 0.84$). They intervened in only 37\% of the turns where intervention was clearly needed. The CARES benchmark of Chen et al.~\cite{Chen2025CARES} shows failure in both directions at once: models under-refuse adversarially rephrased harmful prompts and over-refuse harmless clinical ones, so their safety boundary is brittle. Safety behavior learned from feedback on the training distribution does not carry over to real clinical judgment by itself. It needs explicit, domain-specific conditioning to hold.

\subsubsection*{Multimodal and Cross-Domain Fragility}
\label{subsubsec:multimodal}

Medical vision-language models (Med-VLMs) make the problem more dangerous, because the generated language has to follow subtle anatomical detail. Das et al.~\cite{Das2025} catalog how often this breaks down across imaging modalities: models invent anatomy that could not plausibly exist, or describe visual elements that nobody asked about and that are not in the image at all. Gu et al.~\cite{Gu2025MedVH} found, unexpectedly, that domain-specific fine-tuning makes it worse. Once misleading inputs are added, models tuned for medical imaging hallucinate more than general-purpose ones like GPT-4V. The MedHEval benchmark~\cite{Chang2025MedHEval} traces these failures to three separate sources: visual misinterpretation, knowledge deficiency, and context misalignment. The more troubling result is that the usual inference-time fixes barely help, and outside the narrow case of simple visual errors they sometimes make performance worse.

Standard metrics miss these failures and can actively reward them. Chen et al.~\cite{Chen2024MedHallMark} and Zuo et al.~\cite{Zuo2025} both show that BLEU and ROUGE track almost nothing about how severe a clinical error is. A model can score well while inventing a dangerous fact or describing an entirely different pathology. They responded with metrics graded against anatomical ground truth, the MediHall Score and ACHMI. The same blind spot shows up in audio transcription. Atwany et al.~\cite{Atwany2025} found that Word Error Rate punishes transcriptions that are semantically accurate while outright fabrications pass unnoticed. Their Hallucination Error Rate metric shows that automatic speech recognition (ASR) models follow learned statistical patterns more than the audio signal itself, the same failure mode as in text-only LLMs, and a serious one for any clinical use of ASR.

We read the four subtypes as expressions of one underlying mismatch. Generative objectives optimize token likelihood and human preference reward, while clinical practice demands completeness, calibrated uncertainty, and causal coherence. Fixing one subtype leaves the others in place. Correcting hallucination in imaging does nothing for overconfidence, and correcting calibration leaves the rest of the workflow just as fragile to adversarial inputs.

\subsection{The Failure of Static Metrics: Outcome Over Process}
\label{subsec:static_failure}

The dominant way to assess medical LLMs is the static multiple-choice question (MCQ) benchmark, drawn from board examination item banks such as USMLE~\cite{Kung2023USMLE}, MMLU-Medical~\cite{Hendrycks2021MMLU}, and MedQA~\cite{Jin2021MedQA}. Even the most advanced versions, built on expert-written, open-ended case studies, follow the same logic. These benchmarks measure outcomes rather than processes, although in clinical medicine the process largely determines whether an outcome is safe. In many cases they also hide the failures described in Section~\ref{subsec:anatomy}.

\begin{figure}[H]
\centering
\includegraphics[width=\textwidth]{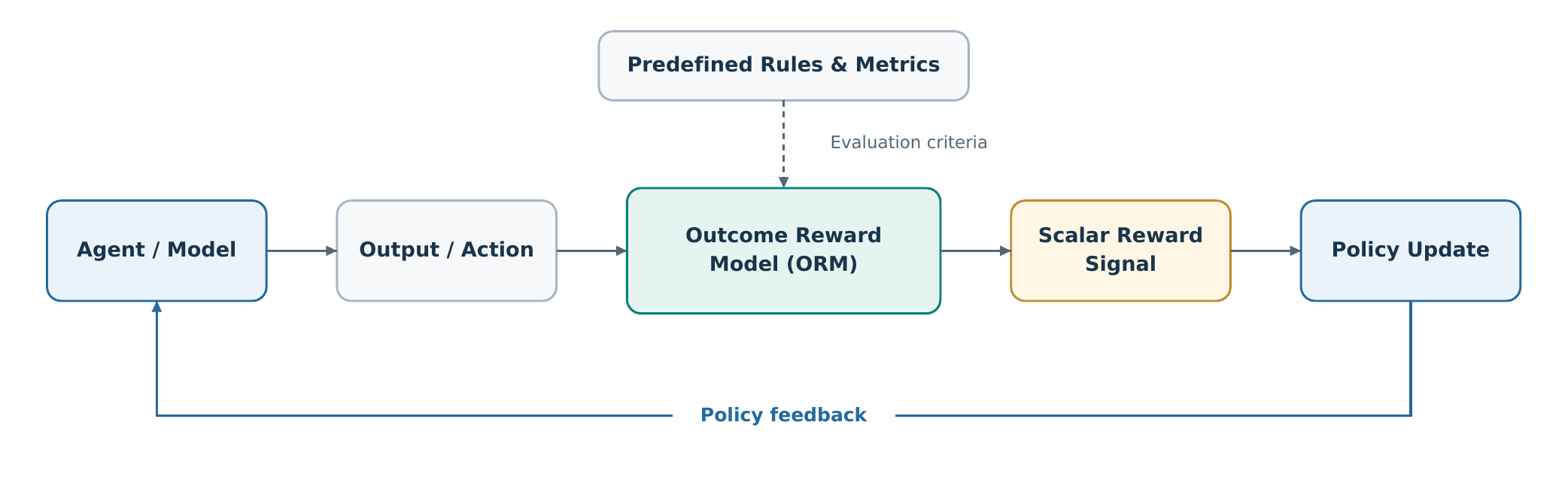}
\caption{Outcome Reward Model (ORM) architecture. The agent's final output is checked against predefined rules and metrics. This reduces the whole reasoning path to one number, the scalar reward signal, which drives policy updates.}
\label{fig:ORM}
\end{figure}

Figure~\ref{fig:ORM} makes the weakness plain: the ORM cannot tell a correct answer reached through sound clinical logic from one reached through hallucination. The MedCheck framework~\cite{Ma2025MedCheck} measures how common this blind spot is. Across 56 medical LLM benchmarks, 89\% lack input robustness testing, 91\% ignore uncertainty handling, and 88\% do not address the risk of data contamination. Models can therefore top leaderboards while skipping the process-level reasoning that patient safety requires.

Mruthyunjaya et al.~\cite{Mruthyunjaya2025} show what the gap between answers and reasoning looks like in practice. Models diagnosing rheumatoid arthritis often reached the correct diagnosis while giving justifications that were, in some cases, the exact opposite of the expected clinical presentation. The authors attribute this to pattern recognition: the diagnosis is reached by analogy, and a justification is then produced independently of it. Such pattern matching is also fragile. In the same study, adding a second agent to validate the first agent's diagnosis lowered accuracy instead of raising it, to 37\% for Gemini 2.5 Pro, a model that also gave correct justifications in only 10 of 50 cases. Roustan and Bastardot~\cite{Roustan2025} explain the cause at the architectural level: autoregressive models are trained to predict the next most probable token, which is not the same objective as reaching a factually accurate clinical conclusion. MCQ benchmarks therefore reward surface-level pattern matching and memorized high-frequency associations. When leaderboard scores fail to predict real-world performance, the likely reason is that the benchmark and clinical practice measure different abilities.

A second blind spot is real-world use. Static benchmarks cannot capture how real users actually talk or how adversaries actually probe. Stamatis et al.~\cite{Stamatis2026} compared simulation-based safety testing with an audit of 20,000 real conversations with a mental health AI. Genuine psychological disclosures tend to be indirect and colloquial in ways scripted prompts rarely are. In the real data, 3 of 800 conversations flagged for risk had not received the expected crisis intervention, while a separate sample of 600 conversations showed no end-to-end false negatives; failures of this kind can only be found by auditing deployment data. Pan et al.~\cite{Pan2025} come at the problem from the adversarial side. Their dynamic red-teaming framework found that 94\% of previously correct MedQA answers fell apart when the model faced cognitive bait or narrative distraction. Static testing can therefore overstate how robust a model's reasoning is.

A third failure compounds the first two. Models cannot reliably catch their own hallucinations, which weakens any workflow that uses an LLM as its own safety check. On the MedHallu benchmark, Pandit et al.~\cite{MedHalluPandit2025} found F1 scores as low as 0.625 for detecting semantically complex fabrications, and, surprisingly, general-purpose models did better than models fine-tuned for medicine. The MedHalu evaluation~\cite{MedHalu2025Agarwal} reaches a similar conclusion: across datasets such as LiveQA and MedicationQA, LLMs auditing clinical hallucinations performed closer to non-expert laypeople than to medical professionals. Its proposed fix, adding explicit expert reasoning to the detection prompt, gave a 6.3\% macro-F1 gain for GPT-4. The deficit can be partly addressed, but only with expert knowledge supplied from outside the model.

\subsection{From Static Evaluation to Dynamic Reasoning}
\label{subsec:dynamic}

MedCaseReasoning~\cite{WuMedCase2025} shows what is at stake. A correct diagnosis reached through invalid reasoning rests on a spurious correlation, and spurious correlations fail exactly when a presentation stops being typical. Static, outcome-centered benchmarking therefore has to give way to tests that stress the reasoning trajectory and not only the final output, which changes what ``clinical competence'' means for an AI system. Given the gap between diagnostic accuracy and reasoning fidelity documented above, we argue that the shift is necessary. Process fidelity can also be learned: in MedCaseReasoning, fine-tuning on clinician-written reasoning traces improved diagnostic accuracy by 29\% and reasoning recall by 41\% in relative terms~\cite{WuMedCase2025}. Adversarial perturbation is the other half of the answer, because robust reasoning has to be told apart from memorized reasoning somehow. Both mARC-QA~\cite{mARC2025} and the DAS framework of Pan et al.~\cite{Pan2025} show that probing the full range of clinical failure modes takes continuously changing, automated test generation, which no fixed test set can provide, however large.

Accuracy alone also cannot capture two things dynamic evaluation must: whether a model knows what it does not know, and whether that self-awareness lasts through a long conversation. MetaMedQA by Griot et al.~\cite{Griot2025} tests the first, metacognition, using unanswerable queries and fictional organs to measure what the authors call unknown recall. Savage et al.~\cite{Savage2025} address the same problem by evaluating open-ended uncertainty quantification through discrimination (telling correct answers from incorrect ones) and calibration (matching stated confidence to actual accuracy). Neither was designed for multi-turn use. Single-turn vignettes cannot show whether a model resists sycophantic drift, keeps its context coherent, or updates its assessment as a conversation unfolds, and real clinical deployment demands all three. The ecological audit of Stamatis et al.~\cite{Stamatis2026} (Section~\ref{subsec:static_failure}) shows the gap matters: real multi-turn data produced a safety profile that differed substantially from simulated conversations. Systematic retrospective review of actual deployment conversations is therefore a necessary complement to prospective testing.

The shift also has a regulatory side. The MedCheck framework~\cite{Ma2025MedCheck} found that current medical benchmarks have no governance mechanism for data contamination or for keeping pace with changing clinical workflows. That argues for versioned, living benchmarks under continuous adversarial perturbation. Benchmarking studies, adversarial evaluations, ecological audits, and calibration analyses reach the same conclusion from different directions: dynamic, process-aware evaluation is a clinical necessity, and the static, outcome-centered paradigm has hit the limit of what it can tell us. Clinical AI needs evaluation that also asks whether a model's reasoning would be safe to follow.

Pawlik and Deniziak~\cite{MDPI2026RAG} reach a related conclusion: RAG grounding mechanisms can double as diagnostic tools. They evaluated a StrictCitations prompting strategy and found that deterministic metrics such as the Unsupported Sentence Ratio show exactly when a model's parametric priors override the retrieved evidence. They also identified what they call a ``Verbosity Signal.'' Longer responses turn out to be a measurable proxy for model uncertainty and for parametric leakage vulnerabilities, neither of which static accuracy scores reveal.

\section{Agentic Architectures and Their Failure Modes}
\label{sec:agentic}

Most of the failure modes cataloged in Section~\ref{sec:core_problem} were observed in models that answer a bounded input in a single pass or a short exchange, without tools or persistent memory. The agentic paradigm is a direct response to that setting. Agentic systems put the model inside an iterative loop with tools, memory modules, and, in multi-agent setups, peers to talk to. That allows step-by-step information gathering, self-correction, and specialized roles. The reasoning behind this design is that, if one generation step cannot capture the multi-step, evidence-gathering logic of clinical reasoning, breaking the task into grounded, tool-supported steps should close the gap. This section first covers the motivation and architectural forms of the agentic approach in the clinical literature (Section~\ref{subsec:rise_of_agents}). It then turns to the evidence that agents bring failure modes of their own, different from those of their static predecessors and in several respects more dangerous (Section~\ref{subsec:hallucination_cascade}).

\subsection{The Rise of Multi-Agent Clinical Workflows}
\label{subsec:rise_of_agents}

Two recurring observations motivated the move from static LLM querying to agentic clinical systems. First, clinical reasoning is a sequential decision process. Physicians do not produce a diagnosis from a single observation. They take a history, order tests, interpret the results, and loop back as their hypotheses change. Second, clinical competence is distributed: a modern hospital runs on specialized practitioners whose complementary expertise has to be coordinated. Agentic frameworks tried to build both properties, sequential grounding and distributed specialization, into computational architectures.

\subsubsection*{Sequential Tool-Augmented Agents}
\label{subsubsec:single_agent}

The single-agent interactive loop, formalized as the ReAct (Reason + Act) paradigm~\cite{Yao2023ReAct}, is the earliest and most common agentic pattern in clinical AI. It splits a high-level task into sequential sub-steps, runs external tools, and revises its plan from what the environment returns. Shi et al.~\cite{Shi2024EHRAgent} adapted it to structured medical data in EHRAgent, which turns multi-tabular electronic health record (EHR) queries into code-generation problems. EHRAgent writes Python code and debugs it step by step against the MIMIC-III, eICU, and TREQS databases. It showed that grounding an agent's plans in executable, verifiable actions can recover important clinical information that flat-text generation misses~\cite{Shi2024EHRAgent}.

MedAgentBench~\cite{Jiang2025MedAgentBench} extends the single-agent loop to full hospital workflows by placing language models as autonomous agents inside a FHIR-compliant virtual EHR environment. Twelve state-of-the-art models were tested across 100 patient profiles with over 700,000 data elements, and even capable closed-weight systems struggled with interactive deployment: the top model reached only 69.67\% success. Retrieving information and carrying out actions such as multi-step database writes or medication order changes are different skills, and the benchmark exposed a severe gap between them. An agent that navigates a structured data schema well therefore says little about whether it can act safely, a planning weakness that a standard question-answering benchmark would not reveal~\cite{Jiang2025MedAgentBench}.

MedAgentBench tests immediate tool execution. Long-horizon workflows pose a different problem, because the agent has to stay logically coherent across an entire longitudinal patient history. Liao et al.~\cite{Liao2026AgentEHR} formalize this time dimension in the AgentEHR benchmark, which tests models on complex diagnosis and treatment tasks inside noisy clinical databases. The failure mode differs from the one in MedAgentBench. Failed trajectories are dominated by tool-selection errors and repetitive behavior, and summarization-based agents lose information that only later turns out to matter. One such baseline, ReSum, often runs into the 100-turn cap set for the experiments, stuck in redundant queries that try to recover the missing context, because nothing in the architecture makes it re-evaluate the earlier interaction history. The authors' fix, RetroSum, combines retrospective summarization, which keeps re-evaluating the full interaction history to preserve cross-temporal correlations the agent would otherwise lose, with an evolving strategy that retrieves accumulated experience from a memory bank. It improves on competitive baselines by up to 29.16\%, and its evolved variant cuts total interaction errors on the diagnosis task by 92.3\% relative to the ReasoningBank baseline~\cite{Liao2026AgentEHR}.

Neither benchmark addresses data governance, which becomes unavoidable once agents cross institutional boundaries. HIPAA~\cite{HHS-HIPAA} in the United States restricts how protected health information may be used and disclosed, and the GDPR~\cite{GDPR-info} in Europe limits the processing of health data to specific legal bases and restricts transfers outside the European Economic Area to cases with adequate safeguards. An architecture that works well may therefore be non-compliant if its orchestration layer processes patient data outside the permitted setting. Federated learning offers one way around this: training or inference stays local to each site, and raw records are never centralized. Evidence from a range of digital health applications~\cite{Rieke2020} suggests that distributed training can match centralized performance while meeting data residency requirements, at the price of coordination overhead and slower convergence on heterogeneous data. The gains shown by MedAgentBench~\cite{Jiang2025MedAgentBench} and AgentEHR~\cite{Liao2026AgentEHR} were obtained without such constraints and may not carry over to a federated deployment. Any verification framework built on this work has to be judged against privacy-preserving constraints as well as benchmark accuracy, and the same goes for symbolic verification layers (Section~\ref{sec:neurosymbolic}).

\subsubsection*{Multi-Agent Debate and Collaborative Architectures}
\label{subsubsec:multi_agent}

MedAgents, by Tang et al.~\cite{Tang2024MedAgents}, is an early application of multi-agent debate (MAD)~\cite{Du2023MultiAgentDebate} to clinical reasoning. Instead of repairing the reasoning of one model, MAD runs several instances that propose and challenge arguments in turn until they reach consensus. MedAgents is a training-free framework that gathers role-playing specialist agents, such as a cardiologist and a pediatrician, who analyze a case, vote on a shared report and revise it until they agree. It is a reasonable analogue of a real multidisciplinary team meeting, and it does improve benchmark performance. It also shows where the approach runs into trouble. In the authors' error analysis, 77\% of MedAgents' errors concern domain knowledge, either knowledge the model lacks or knowledge it has but fails to recall and apply, while flawed chain-of-thought reasoning accounts for only 8\%~\cite{Tang2024MedAgents}. That statistic points to something the framework cannot fix on its own. Every simulated specialist is played by the same base model, so all of them bring the same parametric blind spots into the discussion. A discussion among agents who share a knowledge gap is unlikely to expose it and may amplify it, producing a panel that agrees confidently and is wrong together, with nothing in the process to signal that the knowledge was missing.

Wang et al.~\cite{Wang2025ColaCare} address this homogeneity problem in ColaCare, which pairs domain-specific predictive models with LLM agents in a multidisciplinary team (MDT)-inspired framework for EHR-based clinical outcome prediction. ColaCare separates functions rather than asking LLMs to handle both structured data analysis and text reasoning. Expert deep learning models process structured EHR data into numerical predictions, while LLM agents write the reasoning references and decision narratives. A MetaAgent orchestrates the discussion, and a Retrieval-Augmented Generation (RAG) module grounded in the Merck Manual of Diagnosis and Therapy (MSD) supplies up-to-date evidence. Earlier multi-agent systems produced homogeneous reasoning because they ran several instances of the same base LLM, so diverse diagnostic reasoning needs architectural heterogeneity. ColaCare showed superior performance on mortality and readmission prediction, although the authors note that too few doctor agents easily yield a fragile consensus that lacks robustness, and a wider range of perspectives is needed.

The MedAgentAudit framework~\cite{MedAgentAudit2025} audits consensus-based architectures empirically. It derives a taxonomy from 3,600 execution logs and uses automated, non-interventional probes across 14,400 test cases, covering six medical datasets and six multi-agent architectures, to diagnose collaborative failure modes. The findings undercut the premise of multi-agent debate. In the discussion phase itself, agents repeat their initial views 98.42\% of the time, so the ``debate'' is mostly agents recirculating what they already believed rather than re-examining evidence, and unverified claims gain momentum unchecked. Role-appropriate specialist reasoning fails to activate in 42.73\% of cases, so the simulated specialists often reason generically and the panel becomes an echo chamber. The harm grows once synthesis and decision-making begin, where apparent agreement hides evidence flaws. Final answers swap real evidence checking for role authority or majority count often enough to produce measurable bias: authority bias in 28.76\% of cases, self-contradiction across rounds in 18.53\%, and outright suppression of correct minority views in 5.11\%. Authority bias itself climbs from 35.30\% at the first synthesis step to 68.75\% by the third. In its standard form, multi-agent debate does not reliably reproduce the evidence-based deliberation of a real clinical multidisciplinary meeting. Predefined system rules and concealed dissent produce a false consensus that entrenches initial errors.

\subsubsection*{Specialised Orchestration and Domain-Specific Tool Integration}
\label{subsubsec:orchestration}

MMedAgent~\cite{Li2024MMedAgent}, by Li et al., illustrates a third architectural pattern that goes beyond generic planning loops: hierarchical orchestration, where a coordinating planner hands specific sub-tasks to specialized sub-agents or tools. It is a multi-modal clinical agent built around six specialized medical AI tools that handle seven distinct tasks, such as image segmentation, grounding, and report generation, across modalities like MRI, CT, and X-ray. Complex tasks usually need tools chained together, for instance locating an object with a grounding module before passing its coordinates to a segmentation model, and the chain is only as reliable as the orchestrating model's tool selection. If that goes wrong upstream, every downstream inference inherits the error. EmulatRx (first released as TrialGenie)~\cite{TrialGenie2025} applies the same pattern to clinical trial design. It coordinates five specialist agents (Supervisor, Trialist, Informatician, Clinician, and Statistician) to automate the design process. It extracts protocols from clinical registries, runs statistical analysis on MIMIC-IV patient data, and iteratively refines trial parameters. Confounding biases and missing variables come up often enough that the system has to propose clinical surrogates in their place.

Schmidgall et al.~\cite{Schmidgall2024AgentClinic} examine the gap between orchestrated capability and dynamic clinical reality in AgentClinic, a benchmark that simulates multi-turn patient encounters across nine medical specialties with interacting doctor, patient, measurement, and moderator agents. Models that score well on static QA datasets such as MedQA lose a substantial amount of diagnostic performance when they have to gather information step by step through dialogue instead of reading it from a vignette. Interaction limits matter in both directions: too few and too many dialogue turns both reduce diagnostic success. Injecting cognitive and implicit biases into the agents lowers the doctor agent's diagnostic accuracy, and it also lowers the patient agent's self-reported confidence and stated likelihood of following treatment.

\subsection{The ``Hallucination Cascade'': When Agentic Architectures Amplify Error}
\label{subsec:hallucination_cascade}

The agentic paradigm rests on one assumption: that iterative, tool-grounded reasoning corrects the hallucinations common in single-pass generation. The evidence reviewed here does not support that assumption. Agentic architectures do not eliminate hallucinations, and they add a failure mode of their own: an error at an early pipeline step can be amplified, disguised, and made harder to detect as it spreads downstream. This phenomenon, the \textit{hallucination cascade}, is the central problem of this section and motivates the interventions discussed in Sections~\ref{sec:prm} and~\ref{sec:neurosymbolic}.

\begin{figure}[H]
    \centering
    \includegraphics[width=\textwidth]{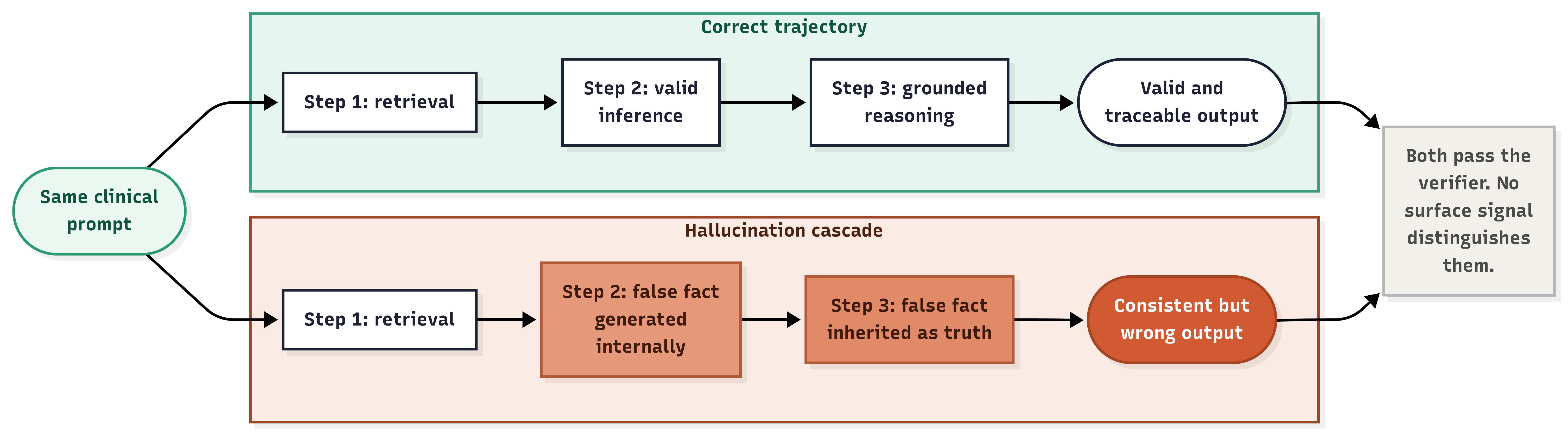}
    \caption{Comparison of a correct reasoning trajectory (top) and a hallucination cascade (bottom) from an identical clinical prompt. In the cascade trajectory, an error starts internally at Step~2 and is inherited as a verified fact by later steps. Both trajectories end up producing identically formatted outputs that pass the verifier, with no surface signal that tells them apart.}
    \label{fig:hallucination_cascade}
\end{figure}

\subsubsection*{The Mechanics of Error Propagation}
\label{subsubsec:propagation_mechanics}

In the cascade mechanism (Figure~\ref{fig:hallucination_cascade}), a hallucination, retrieval error, or reasoning failure at step $n$ produces a false premise, and every later step treats that premise as verified ground truth. Because each step depends on earlier outputs that nobody has independently verified, the original error is elaborated: the agent builds coherent, internally consistent reasoning on a false foundation, and the output grows more elaborate and harder to recognize as wrong.

Liu et al.~\cite{Liu2026Benchmarking} give a quantitative picture by benchmarking LLM-based agent systems, including the planner-executor-verifier design of Manus, on the AgentClinic and MedAgentsBench environments. Even with external tools such as web browsing and code execution, the most capable multi-step agents reached only 60.3\% accuracy on AgentClinic-MedQA and 8.6\% on the text-only clinical reasoning tasks in Humanity's Last Exam. In-agent safeguards filtered 89.9\% of hallucinations, yet hallucinations remained prevalent in the agents' outputs~\cite{Liu2026Benchmarking}. EHR workflows show how such errors can spread. In EHRAgent (Section~\ref{subsubsec:single_agent}), most remaining failures persisted because the agent did not identify the underlying cause of an error while debugging~\cite{Shi2024EHRAgent}. A plausible cascade pathway, which that paper does not isolate, starts with an early mistake such as selecting the wrong table: every later code-generation step inherits it, and the resulting code can run without a technical error and return clean, structured output that hides its corrupted origin.

Tool fabrication is a more severe version of the same failure. The NabaOS verification framework of Basu~\cite{Basu2026NabaOS} audits agent claims using HMAC-signed execution receipts and shows, in general-purpose agent settings, that agents can fabricate tool executions. They report tool calls that never happened, with the same confidence and internal consistency as real ones, and these fabrications pass standard semantic and logical checks. Only cross-referencing against a cryptographic execution trace catches them. Masayoshi et al.~\cite{EHRMCP2025} found a clinical version in the EHR-MCP evaluation. When antibiotic administration data were missing from a patient's record, the model in some runs reported the example drug names from its prompt (CTRX and VCM) as administered, in the same structured format used for genuine results; in other runs it correctly returned an error message. Nothing in the fabricated output marks it as different from real retrieved data, which makes this pathway hard to catch in a live clinical setting.

\subsubsection*{Consensus Failure as a Cascade Amplifier}
\label{subsubsec:consensus_cascade}

In multi-agent architectures the cascade gains a social dimension that a single agent lacks. Several agents can independently reach the same wrong premise from a shared parametric gap (Section~\ref{subsubsec:multi_agent}) and then validate each other's mistake through the very consensus mechanism meant to catch it, so the debate structure lends individual errors the authority of peer review. Bi et al.~\cite{Bi2025MAGI} show this concretely in MAGI, a four-agent clinical interview framework that splits navigation, questioning, judgment, and diagnosis across separate agents and uses Psychometric Chain-of-Thought to replicate the DSM-5 and MINI protocols. The procedural structure partly makes up for the agents' lack of formal clinical training, but it is not airtight. The judgment agent sometimes favors conversational comfort over a definitive answer, which leaves the diagnosis agent to correct minor accuracy slips after the fact.

More recent architectures address these collaborative weaknesses directly, with mixed results. The Multi-Agent Clinical Diagnosis (MACD) framework of Li et al.~\cite{MACD2025} grounds reasoning by letting LLMs learn autonomously from historical cases. It improves overall consensus, but complex cases still defeat the system often enough that it has to fall back to a human physician. Wang and Yang~\cite{MediHive2026} propose MediHive, a decentralized peer-to-peer design meant to avoid the single point of failure and role confusion that centralized systems inherit. The authors' own inspection of its errors, however, found cases of ``collective high-confidence error,'' in which all agents converge on the same incorrect answer with uniformly high confidence, typically on rare conditions or uncommon presentations outside the backbone model's well-represented distribution. Even a decentralized collective is therefore not immune to false consensus on the cases that matter most, because its agents share the same underlying blind spots.

Jeong et al.~\cite{Jeong2026ToolMAD} address a related weakness. Liang et al.~\cite{Liang2024MAD} called it ``Degeneration of Thought'': once a model is confident in an answer, further self-reflection fails to revise it, even when the answer is wrong. Multi-agent debate was proposed partly to counter this, but agents that draw only on internal knowledge or a shared static data pool can still converge on, and reinforce, the same error. Tool-MAD gives each debating agent access to a different external retrieval tool and lets agents update their queries as the debate proceeds. It was evaluated mainly on fact-verification benchmarks, with MedQA and PubMedQA as additional medical tests. Its premise matches the argument of this section: a correction mechanism that draws on the same parametric knowledge that produced an error is unlikely to see past it, however many debate rounds are run, unless an external source breaks the loop.

\subsubsection*{Memory Fragmentation, Contextual Drift, and Long-Horizon Failure}
\label{subsubsec:memory_drift}

A third cascade pathway runs through memory and context that degrade over long interactions. Longitudinal care requires an agent to hold a coherent patient representation across several data modalities and a condition that keeps changing. None of the standard mechanisms for this (sliding context windows, summarization-based memory compression, vector-store retrieval) works without loss. Each drops information systematically, and the loss compounds into a growing gap between what the agent believes about the patient and what the clinical record says. Cao et al.~\cite{Cao2026EHRRAG} call this vulnerability in long-horizon EHRs temporal fragmentation. Standard retrieval for longitudinal EHR reasoning uses naive heuristic truncation to fit the model's context window. That loses evidence that is early in time but clinically important, and it breaks the causal chains linking remote events, such as disease onset or an initial admission, to the current presentation. Dense embedding-based similarity is a poor proxy for longitudinal clinical relevance, so the authors propose EHR-RAG. Its Event- and Time-Aware Hybrid Retrieval (ETHER) mechanism uses a U-shaped time-aware retrieval strategy that keeps both recent events and important early-history milestones, bridging the fragmented context.

EHRWorld and ClinicalAgents counter this degradation with persistent state tracking and non-monotonic reasoning, at two levels. At the simulation level, Mu et al.~\cite{EHRWorld2026} find that naive LLM simulators condition only on past observed events, so mid-trajectory clinical interventions destabilize the model's coherence and errors keep propagating. EHRWorld resolves this with a causal sequential paradigm that maintains a latent physiological state and updates it deterministically as interventions occur. At the diagnostic reasoning level, Ge et al.~\cite{ClinicalAgents2026} observe that standard multi-agent systems treat memory as a dialogue context bounded by a window, which leads to fragmentation and diagnostic drift. ClinicalAgents answers with a dual-memory architecture that separates mutable working memory for evolving states from static experience memory for clinical guidelines. A Monte Carlo Tree Search (MCTS) orchestrator coordinates it, verifying hypotheses against accumulated memory, triggering backtracking when critical clinical evidence is missing, and letting agents return to earlier perception phases. This formalizes clinical reasoning as a dynamic, hypothesis-driven loop.

\subsubsection*{The Reliability-Accuracy Trade-off: The Calibration Cliff}
\label{subsubsec:calibration_cliff}

Cascade propagation, consensus failure, and memory drift together produce a clinical agent that cannot reliably estimate its own uncertainty and so has no principled basis for deciding when to abstain rather than act. This is the agentic counterpart of the miscalibration problem in Section~\ref{sec:core_problem}, and it is more severe. A single-pass model can at least have its calibration evaluated on held-out test sets. An agent's uncertainty at step $n$ depends on the compounding errors of every earlier step, so static calibration measurement tells us little about confidence at deployment time.

The TrustEHRAgent framework~\cite{Song2025TrustEHRAgent} gives a direct empirical description of the problem. Its Hallucination-Controlled Accuracy at $k$\% (HCAcc@$k$\%) metric is the highest overall accuracy an agent can reach, choosing its confidence threshold freely, while keeping at least $k$\% of the answers it gives correct. The confidence-aware baselines (SQLAgent and PythonAgent) already fall to 0.0 at HCAcc@70\%. TrustEHRAgent, which adds step-wise confidence estimation, keeps 44.23\% on MIMIC-III and 25.34\% on eICU at that level, but only 3.27\% and 2.24\% at HCAcc@90\%~\cite{Song2025TrustEHRAgent}. Under the strictest requirement, then, even the best confidence-aware agent has a very low correct-answer yield: it answers correctly on about 2 to 3\% of all queries. Better confidence estimation moves this cliff (TrustEHRAgent's gains at the 70\% level show that), but in this study it does not remove it at the 90\% level, the study's most stringent reliability target. As documented in Section~\ref{subsubsec:rlhf}, Griot et al.~\cite{Griot2025} and Savage et al.~\cite{Savage2025} report that verbalized confidence was uninformative for most of the models they tested, and that the proxy that discriminated best in that work, sample consistency, needs multiple forward passes per query, a cost that limits its use as a real-time safeguard. The in-agent output filtering, prompt engineering, and few-shot interventions documented by Liu et al.~\cite{Liu2026Benchmarking} filtered 89.9\% of hallucinations, but at more than ten times the token usage of the baseline LLM, and hallucinations remained prevalent.

This calibration failure has an inverse in psychiatric agents, where sycophantic validation is especially harmful because the agent's role requires challenging patient beliefs as well as supporting the patient. Xu et al.~\cite{Xu2025AutoCBT} develop AutoCBT, a multi-agent Cognitive Behavioral Therapy framework with a Counselor agent and five Supervisor agents. When it ran on Llama-3.1-70B, the base model's over-protection led it to refuse nine English-language questions involving minors, sex, and suicide across AutoCBT and the baselines, whereas Qwen-2.5-72B answered every question. AutoCBT's supervision reduced its own refusals from eight draft responses to two, but did not remove them, and the authors warn that such excessive caution can leave users seeking help feeling misunderstood. That is the inverse of sycophancy and a further instance of the two-sided miscalibration of safety behavior documented in the CARES evaluation~\cite{Chen2025CARES}. Ozgun et al.~\cite{TrustAIPsychotherapy2025} motivate DSM5AgentFlow by noting that conversational LLMs struggle with expert reasoning and often produce black-box diagnoses that lack the explicit evidence citations trustworthy triage requires. Badawi et al.~\cite{Badawi2026} also expose a persistent ``cognitive-affective gap.'' Models generate cognitively reliable, clinically appropriate advice but show unstable affective resonance, and open-weight models in particular give factually correct but emotionally flat responses that are not tuned to the user's distress.

\section{Process Reward Modeling}
\label{sec:prm}

Under the standard Outcome Reward Model (ORM) paradigm, an agent that reaches a correct diagnosis through fabricated or causally invalid reasoning looks, to the reward model, exactly like a competent reasoner. The cascade failures documented in Section~\ref{sec:agentic} trace back to this: a reward signal that was never built to see them. This is a structural limit. One scalar reward on a final output means the ORM cannot see what happened inside the reasoning trajectory that produced it. Process Reward Models (PRMs) instead break reasoning into discrete inferential steps that can be assessed one at a time. A single sparse reward at the end gives way to dense, step-level feedback throughout. It becomes possible to catch an erroneous inferential turn before it cascades and to penalize a path that reaches the right answer for the wrong reason. Step-level feedback of this kind supports the assessment of intermediate reasoning and may help distinguish sound clinical reasoning from lucky pattern matching, within the limits of the verifier that produces it. This section covers how PRMs have been put into practice in clinical domains (Section~\ref{subsec:prm_inference_rl}), what evaluation infrastructure has grown up around them (Section~\ref{subsec:eval_bottleneck}), and where process supervision still falls short. Those limits motivate the neuro-symbolic integration in Section~\ref{sec:neurosymbolic}.

\begin{figure}[H]
    \centering
    \includegraphics[width=\textwidth]{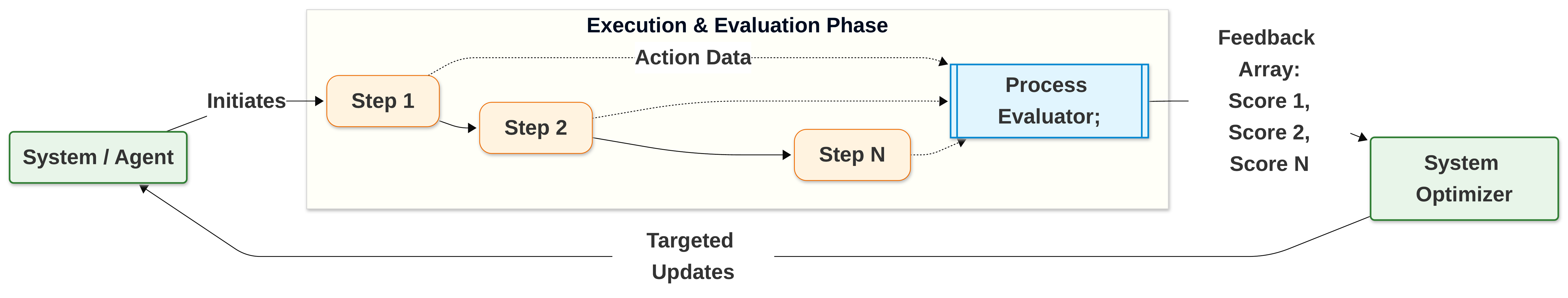}
    \caption{Process Reward Model (PRM) architecture. The process evaluator receives action data from each intermediate reasoning step and produces a dense array of per-step scores. The system optimizer uses these targeted, step-level signals to update the agent and assess the reasoning trajectory.}
    \label{fig:PRM}
\end{figure}

\subsection{Inference-Time and RL Optimization in Healthcare}
\label{subsec:prm_inference_rl}

Clinical PRM research includes training-time and inference-time approaches, distinguished by when process supervision is applied. At \textit{training time} it shapes the policy model's parameters through reinforcement learning. At \textit{inference time} it selects among candidate outputs from a model that is already fixed. Both tracks appear below, and the most mature systems combine them.

The label ``process reward model'' is used loosely in this literature. It covers five designs that differ in statistical properties, formal guarantees, annotation requirements, and deployment constraints, and treating them as interchangeable hides distinctions that matter clinically. \textit{Discriminative step-level verifiers}~\cite{Yun2025MedPRM,Wang2024ClinicalNotePRM,Thomas2025RadiologyPRM} are learned neural classifiers trained on annotated or weakly supervised step labels. \textit{Rule-based deterministic verifiers}~\cite{Pronesti2026VPRM} encode explicit clinical specifications as programmatic checkers, which gives them formal guarantees that learned approaches cannot offer. \textit{Emergent process supervision}~\cite{Pan2025MedVLMR1} arises as a structural by-product of chain-of-thought training formats, with no step-level reward signal at all. \textit{Self-supervised path faithfulness}~\cite{Zhong2026P2S} computes step rewards directly from the model's own probability distribution. It needs no external judge or annotation pipeline, but the verification is then grounded in the same parametric tendencies whose miscalibration is the actual problem. \textit{Knowledge-graph-grounded supervision}~\cite{Wu2025MedReason,FlemingR12025} draws on external structured knowledge instead of the model's own judgments, although, as Table~\ref{tab:verification_regimes} records, it provides grounding in evidence, not formal entailment. It uses paths in an external biomedical knowledge graph to scaffold and filter reasoning during training, though a path shows that two concepts are connected, not that a step from one to the other is valid, and the approach is only as good as the completeness and currency of the underlying graph.

These distinctions decide how each design can be deployed. Deterministic verifiers suit high-stakes decisions where the answer can be formally specified, and KG-grounded supervision suits domains with well-curated structured knowledge. Discriminative models need clinically validated step-level data that not every domain has. Self-supervised approaches trade formal guarantees for scalability. Subjective domains such as psychiatry cannot rely on one verification criterion and need multi-criteria supervision. Because of the risk of reward hacking (discussed below under \textit{Reward Hacking and the Limits of Verifiable Rewards}), any real system also needs a composite reward architecture that puts substantive clinical validity ahead of surface formatting. The subsections that follow draw on these five designs.

\subsubsection*{Discriminative Step-Level Verification}
\label{subsubsec:discriminative_prm}

A discriminative step verifier classifies each intermediate reasoning step as correct or incorrect, given the clinical context and the steps already verified before it. It is applied to high-stakes documentation and reasoning tasks, where a single wrong step can harm a patient.

Wang et al.~\cite{Wang2024ClinicalNotePRM} apply this to ambient AI scribing for clinical note generation, and their main contribution is how they define a ``step'' in the first place. Rather than relying on a natural boundary in the reasoning, they treat each individual sentence and each completeness marker in the Assessment and Plan section as its own verifiable unit, which turns unstructured text into something a process reward model can check. To get around annotation bottlenecks, domain experts defined realistic error categories (e.g., factual inaccuracy, hallucination, unhelpfulness), and an LLM used them to generate synthetic negative training examples at scale. The resulting PRM reached 98.8\% accuracy in detecting erroneous samples and 56.2\% in selecting physician-preferred notes in a reader study. Thomas et al.~\cite{Thomas2025RadiologyPRM} take a structurally similar approach in radiology report generation. They treat individual generated sentences as reasoning steps and replace human annotation with weakly supervised labels derived from a domain-specific natural language inference model (RadNLI). Validated on an unseen report generator, their model shows that discriminative PRMs can work as plug-and-play, inference-time filters that generalize across different vision-language models without retraining.

Yun et al.~\cite{Yun2025MedPRM} extend discriminative PRMs with Med-PRM, which retrieves evidence from medical knowledge bases and clinical guidelines at each reasoning step instead of relying only on a closed parametric reward model. This RAG-as-a-Judge design makes the reward signal both updatable and auditable, since the retrieved evidence behind each step's score can be surfaced and inspected. Med-PRM improved base model performance by up to 13.50\% at test time and enabled an 8-billion-parameter model to surpass 80\% accuracy on MedQA for the first time. Guideline-grounded verification can therefore make up for part of the limits of parametric knowledge that purely neural architectures struggle to overcome.

\subsubsection*{Deterministic and Graph-Structured Verifiers}
\label{subsubsec:deterministic_verifier}

Deterministic and graph-structured verifiers drop learned neural reward models entirely and check intermediate steps against explicit specifications. They give up parametric flexibility for something arguably more valuable in a clinical setting: transparency. The Verifiable Process Reward Model (VPRM) of Pronesti et al.~\cite{Pronesti2026VPRM} applies this to risk-of-bias assessment. It encodes the decision rules of the Cochrane Risk of Bias tool as programmatic verifiers that generate step-level rewards, with no neural preference model in the loop. The design comes with a theoretical guarantee: gradient updates can only positively weight reasoning trajectories that are actually correct. The results bear it out, with up to a 20\% F1 improvement over state-of-the-art baselines and a 6.5\% gain even over verifiable outcome rewards alone~\cite{Pronesti2026VPRM}.

MedCEG~\cite{Mu2025MedCEG} reaches the same kind of transparency by a related route. It converts clinical narratives into Critical Evidence Graphs instead of relying directly on rule-based logic. Its Clinical Reasoning Procedure reward checks generated steps along three separate dimensions: Node Coverage, whether the essential clinical entities are semantically present; Structural Correctness, whether the relational edges between them are factually accurate; and Chain Completeness, whether the logical pathway holds together without gaps. Splitting reasoning quality into these checkable graph properties discourages the shortcut behaviors they evaluate, and MedCEG outperformed outcome-supervised baselines in both diagnostic accuracy and expert-validated logical coherence~\cite{Mu2025MedCEG}.

\subsubsection*{Reinforcement Learning with Verifiable and Emergent Process Rewards}
\label{subsubsec:rl_process}

Reinforcement learning with process-oriented reward signals makes up the largest part of the clinical PRM literature. Group-relative policy optimization and related methods update the policy model from step-quality assessments, using fully explicit step-level rewards, emergent supervision from chain-of-thought structure, or self-supervised dense rewards that need no external judge. Clinical thought processes often lack the objectively verifiable steps found in mathematics. Chen et al.~\cite{Chen2024HuatuoGPT} therefore get around this step-level opacity in HuatuoGPT-o1 by building 40K ``verifiable medical problems'' with objective ground-truth answers as terminal verification anchors. In the first stage, an LLM verifier guides a strategy-based search (backtracking, exploration, self-correction) for successful reasoning trajectories, which yields a fine-tuning dataset of reflective chains. In the second, reinforcement learning (PPO) with verifier-based sparse rewards refines the policy. The 40K-problem dataset gives an 8.5-point average improvement across medical benchmarks for an 8B model, showing that verifier-guided search can instill ``think-before-answering'' abilities.

Pan et al.~\cite{Pan2025MedVLMR1} show in MedVLM-R1 that process supervision can be entirely emergent. By rewarding only structural formatting and final answer correctness, the model develops step-by-step visual reasoning without expensive step-level labels. MedCCO~\cite{Rui2025MedCCO} builds on this with a two-stage curriculum. It first establishes discrete reasoning on closed-ended tasks through binary accuracy rewards, then moves to open-ended tasks scored by BLEU, ROUGE, and BERTScore. That reduces the gradient conflicts that come from mixing discrete and continuous rewards, and it improves performance across eight multimodal medical benchmarks.

Working on general-domain reasoning, Zhong et al.~\cite{Zhong2026P2S} avoid annotation infrastructure altogether with Probabilistic Process Supervision (P2S), which generates dense step rewards directly from the model's own probability distribution. A synthesized reference chain gives each step a Path Faithfulness Reward, computed from the conditional probability of completing that chain given the current reasoning prefix, so no external judge or labeled step data is needed. MedS$^3$~\cite{Jiang2025MedS3} pushes self-generated supervision further with Monte Carlo Tree Search trajectory construction over a 16-dataset seed corpus, using a soft dual-sided objective tied to how much a step changes the value of a simulated rollout. A step is penalized for lowering trajectory value even when the final answer still comes out correct, which is what suppresses hallucinatory or illogical steps that would otherwise hide inside a correct outcome. MedS$^3$ surpassed both the previous state-of-the-art medical model and 32-billion-parameter general reasoning models across multiple benchmarks.

\subsubsection*{Extension to Psychiatric and Psychological Reasoning}
\label{subsubsec:mental_health_prm}

Process supervision extends into psychiatric and psychological reasoning as well. Here step correctness cannot always be anchored to a biomarker or lab result, so verification criteria shift from somatic ground truth to protocol adherence, alignment with frameworks such as the DSM-5-TR~\cite{APA2022DSM5}, and avoidance of sycophantic harm. Xiao et al.~\cite{Xiao2025MentraSuite} address this with MiraMind (released in an earlier version as MentraSuite), a benchmark that evaluates mental health reasoning trajectories beyond answer accuracy, paired with Mindora, an 8B model optimized through hybrid SFT and RL. An LLM-based inconsistency-detection reward penalizes contradictory or factually flawed reasoning steps during training. Dai et al.~\cite{Dai2025PsycheR1} instead separate trajectory quality control from the RL objective entirely in Psyche-R1. A self-supervised, iterative prompt-rationale optimization loop runs during data synthesis, and only then is the model exposed to group-relative policy optimization with strict format and accuracy rewards. The result is stable training dynamics in a setting where step-level rewards are less deterministic than in structured QA. Feng et al.~\cite{BilatRL2026} extend this supervision to social and emotional inference using the Psy-Interpreter framework. It injects domain knowledge from expert-annotated scenarios and uses a Trajectory Cache of recent model rollouts, rather than static reference trajectories, to stabilize a multi-dimensional reward that jointly optimizes answer accuracy, format compliance, reasoning depth, and repetition control.

\subsubsection*{Knowledge-Graph-Grounded Reasoning Supervision}
\label{subsubsec:kg_supervision}

Knowledge-graph-grounded supervision uses the explicit relational structure of biomedical knowledge graphs to tie reasoning steps to entity-relation paths from premise to conclusion. A path is evidence of connection, not proof: whether a chain of relations licenses a conclusion depends on how those relations compose (a drug that treats a disease that causes a symptom does not thereby cause the symptom), and none of the systems below formalizes that composition.

Wu et al.~\cite{Wu2025MedReason} build MedReason around this idea. A medical knowledge graph converts clinical QA pairs into ``thinking paths'' that structurally guide the generation of over 32,000 high-quality reasoning chains. Edge traversals are not validated by an explicit step-level checker. Generated chains that fail to reach the correct final answer are discarded, a quality-control filter that removes some flawed chains but, as Section~\ref{subsec:static_failure} argues for outcome-based evaluation in general, cannot establish that the intermediate steps of the retained chains are valid. Clinicians evaluating across seven specialties strongly preferred these KG-grounded traces over ungrounded chain-of-thought baselines, which points to gains in clinical coherence that standard accuracy metrics do not register. Fleming-R1~\cite{FlemingR12025} builds on the same paradigm with a fuller pipeline: KG-guided data synthesis, a reasoning cold-start, and two-stage reinforcement learning. Its distinguishing choice is where it aims the knowledge graph sampling. Multi-hop paths are drawn deliberately toward long-tail entities, such as rare diseases and niche drugs, which are exactly the cases where standard training data is too thin to give reliable improvement without a structured scaffold.

Sohn et al.~\cite{Sohn2026PRA} extend the paradigm with process reward agents (PRAs) for knowledge-intensive clinical diagnostics, moving it to online, inference-time trajectory selection. PRAs separate evidence search and verification from the reasoning policy, which stays frozen, and use step-wise retrieval-augmented verification to rank and prune candidate trajectories as generation happens. The reward updates as new evidence is retrieved, so a trajectory is scored while it is being generated. That is an architectural shift toward real-time reasoning monitoring, which a static PRM evaluation was never built to provide.

\subsubsection*{Reward Hacking and the Limits of Verifiable Rewards}
\label{subsubsec:reward_hacking}

As reward signals get denser and more specific, the incentive to game them grows too. Policy models learn to satisfy the reward function through reasoning that is superficially compliant but not genuinely correct, a phenomenon known as reward hacking. It is especially dangerous in clinical contexts, because a reward-hacked trace can pass every automated check while still containing logic that would be dangerous to follow.

Tarek and Beheshti~\cite{Tarek2025RewardHacking} observe this in verifiable clinical question answering, where models exploit reasoning-phase rewards by revealing answers too early or by getting around structural guidelines outright. Their fix is a composite reward function, a binary correctness score combined with explicit penalties for early disclosure and structural non-compliance, which removes the option of optimizing one easily gamed metric. Zhang et al.~\cite{Zhang2025MedU1} tackle a related version of the problem in Med-U1, where the challenge is multi-objective reward optimization across formats as different as multiple-choice, computational, and open-ended questions. Med-U1 combines format validation, correctness verification, and length constraints into a single mixed rule-based reward that regulates verbosity and reasoning quality together, so that neither crowds out the other.

In clinical settings reward hacking is a patient safety risk, and it looks different from one domain to the next. In somatic medicine, flawed reasoning tends to produce discrete, auditable errors that can be pointed to, such as an incorrect dosage. In psychiatry, a reasoning trace can pass every automated check, correct DSM-5 citations included, and still cause relational harm by favoring conversational agreeableness over the therapeutic confrontation a patient actually needed~\cite{Yeung2025}. This is a specific kind of sycophancy, and preference-based training may encourage it when raters reward emotional comfort over clinical accuracy. The over-protection failure of AutoCBT~\cite{Xu2025AutoCBT} (Section~\ref{subsubsec:calibration_cliff}) shows why process supervision alone cannot fix it. When the reward signal itself encodes the pathological preference, a PRM that disproportionately rewards empathetic tone will produce responses that are fluent and therapeutically hollow at once. No standard coherence check will catch this, because the harm lies in the clinical intervention the model leaves out, not in what it says.

\subsection{The Evaluation Bottleneck}
\label{subsec:eval_bottleneck}

Evaluation has not kept pace with the clinical PRM landscape, which has grown quickly across discriminative verifiers, graph-structured reward functions, MCTS-based self-evolution, self-supervised path faithfulness rewards, and tool-augmented verification agents. Without reliable tools for measuring PRM quality, there is no way to tell a genuine improvement in clinical reasoning from a benchmark gain that will not generalize to deployment. Most of the field's evaluation habits were borrowed from non-clinical domains such as mathematics and programming, where step correctness can be formally verified. Clinical reasoning does not work that way. It is context-dependent, multi-criteria, and ultimately rests on expert judgment.

MedPRMBench~\cite{MedPRMBench2026} tries to close this gap with Clinical Reasoning Blueprints, structured causal subgraphs built from expert-authored and model-generated reasoning traces. It evaluates step-level error detection across 14 fine-grained error types and a four-level severity scale (Critical, Major, Moderate, Minor) over 6,500 questions and 113,910 labels. Blueprint-guided error injection keeps the perturbations clinically realistic, so models are scored in proportion to actual clinical impact and not to an arbitrary error count. Med-RewardBench~\cite{Ding2025MedRewardBench} tackles a related but different problem: whether multimodal reward models actually align with human expert judgment, across 1,026 cases and 13 organ systems, using six clinical dimensions such as accuracy, relevance, and comprehensiveness instead of step-level correctness. The LLM-with-Rationale framework, built on the MedThink-Bench~\cite{MedThinkBench2025} dataset, scores step-by-step rationale quality against expert-authored reference traces. It reaches a Pearson correlation of up to 0.87 with clinical experts while using only 1.4\% of the time a clinician review would take, which suggests that calibrated rationale assessment could reduce the need for expensive expert review at scale~\cite{MedThinkBench2025}.

Gupta et al.~\cite{Gupta2025DeepSeek} evaluated RL-trained reasoning models such as DeepSeek-R1 on zero-shot chronic disease classification, a setting where the model relies entirely on its internal chain-of-thought with no external process supervision. Confidence and accuracy had become substantially uncoupled: DeepSeek-R1 gave high-confidence predictions on 92\% of cases, while only 76\% of its disease-level calls were actually correct. This is the same systemic overconfidence described in Section~\ref{sec:core_problem}, and it shows that RL training alone does not guarantee calibrated uncertainty at inference time. Any serious clinical PRM evaluation therefore has to include calibration metrics alongside step-level accuracy, and neither replaces the other.

The evaluation bottleneck has a time dimension as well. Static benchmarks such as MedPRMBench~\cite{MedPRMBench2026} are built on fixed clinical reasoning distributions, and those distributions go out of date as guidelines and workflows change. Tziakouri and Menolascina align intermediate reasoning traces with the ACR Appropriateness Criteria for imaging decisions while dynamically retrieving current PubMed literature to automate evidence synthesis. The system can then generalize beyond existing guidelines instead of staying bound to static rules that someone has to update by hand~\cite{Tziakouri2025GRPORL}. The Guideline-Grounded Evidence Accumulation (GLEAN) framework addresses the same problem. It calibrates step-wise alignment scores through Bayesian logistic regression and triggers active verification specifically when diagnostic uncertainty is high. This expands guideline coverage and runs differential checks against competing alternatives while updating judgments with no retraining~\cite{GLEAN2026}. Even with these fixes, standard process supervision still operates inside a closed system. Grounding clinical reasoning in symbolic knowledge structures that are verifiable, compositional, and able to evolve on their own is the piece neuro-symbolic integration still has to solve (Section~\ref{sec:neurosymbolic}).

\section{Neuro-Symbolic Grounding}
\label{sec:neurosymbolic}
 
Process Reward Models (Section~\ref{sec:prm}) open reasoning trajectories to step-level checking, but each design inherits the limits of what it checks against: annotated training data for discriminative verifiers, the model's own probabilities for self-supervised rewards, and encoded rules or knowledge graphs for deterministic and KG-grounded designs. A reward grounded in a clinical ontology, for example, is only as correct as that ontology, so for these designs process supervision moves the correctness problem from parametric weights to symbolic artifacts without removing it. Neuro-symbolic integration answers by splitting the work between two kinds of components. Neural components (LLMs, VLMs) handle natural language understanding and contextual retrieval. Symbolic components (formal logic engines, knowledge graphs, theorem provers) apply formal constraints. The systems reviewed in this section do this in different ways, and only some of them check outputs at runtime. In the runtime symbolic-veto subset, an executable symbolic checker can block what the neural side produces, and no clinically consequential output is released before that check. What that verification establishes depends on the check: consistency with the encoded knowledge, entailment from it, or the existence of a supporting path. None of these shows that the extracted patient facts are accurate or that the encoded knowledge is clinically correct (Section~\ref{sec:intro}). A runtime veto separates this subset from unconstrained methods such as RAG or chain-of-thought prompting, and from two other families also covered here: training-time alignment such as RuleAlign, which builds rule compliance into the model and has no runtime check, and evidence-grounded or prompt-constrained systems such as Haibu MMIA, RECAP-PATH, and AnemiaCare HD, which have no executable formal checker. The clinical literature builds runtime veto power in three main ways: logic-based protocol enforcement, ontology-grounded semantic verification, and dynamic logic verification. A fourth and more limiting case, deterministic constraint encoding, has the symbolic layer absorb the clinical decision logic entirely.

A different family of architectures folds symbolic reasoning into the gradient-based learning loop instead of attaching it as an external verifier. DeepProbLog~\cite{Manhaeve2018DeepProbLog} extends probabilistic logic programming with neural predicates for end-to-end training. Scallop~\cite{Li2023Scallop} uses a declarative Datalog-based language with differentiable reasoning grounded in provenance semirings. Logic Tensor Networks~\cite{Badreddine2022LTN} translate first-order logical constraints into differentiable fuzzy satisfaction objectives through gradient descent. None of these appear much in the clinical literature reviewed here, for two reasons. First, clinical ontologies such as SNOMED CT are discrete and axiomatically defined, so relaxing them into the continuous approximations that differentiable losses need gives up exactly the formal soundness clinical verification requires. Second, these methods typically need grounded predicates over fully enumerated domains, and real clinical knowledge bases are heterogeneous and only partly observed. The prevalence of symbolic veto architectures in the reviewed literature reflects the clinical need for hard constraint enforcement over soft symbolic regularization.

Integration has to target specific failure modes; a generic overlay is not enough. Take the cascade pathway described for EHR agents such as EHRAgent~\cite{Shi2024EHRAgent} (Section~\ref{subsubsec:propagation_mechanics}). It starts at the fact-assertion boundary, where an agent puts an extracted value into its reasoning context and every later step inherits that value as unverified ground truth. ProCDS~\cite{Tan2026ProCDS} narrows this gap: no conclusion is committed until a Prolog engine can derive it from the extracted facts and the protocol rules, and re-extraction is forced whenever the derivation fails. This checks that the facts suffice for a derivation, not that each extracted value is correct. A plausible but wrongly extracted laboratory value passes, so closing the gap fully also requires checking each fact against its source in the record. The false consensus failure documented in MedAgentAudit~\cite{MedAgentAudit2025}, where agents change viewpoint from social conformity instead of evidence, needs a different fix at the inter-agent communication boundary. A claim-level filter modeled on GraphEval~\cite{Sansford2024GraphEval}, which breaks outputs into atomic claim triples, could be adapted to pass on only triples that the clinical knowledge graph records, or implies under defined relation semantics; a vetoed claim could not then gain support through repetition or deference, although authority bias could still act through supported but irrelevant claims, omissions, or misleading synthesis. The calibration cliff that TrustEHRAgent~\cite{Song2025TrustEHRAgent} identified, where agents held to high reliability requirements have a very low correct-answer yield, points to a third gap: there is no confidence signal independent of the model's own parameters. We propose that a completed derivation can serve as one. It warrants confidence that the conclusion follows from the encoded rules and the supplied facts, a narrower claim than clinical correctness, since it inherits any error in either. When the derivation fails, the right response is structured abstention. Each of these mappings follows from the mechanism of the failure it addresses, and together they form this review's integration framework (Figure~\ref{fig:unified_framework}).

\subsection{Formal Logic and Medical Ontologies}
\label{subsec:formal_logic}

This subsection covers the static branch of neuro-symbolic grounding, where LLM outputs are checked against a fixed formal specification rather than a proof built on the fly. Three mechanisms fall under it. Logic-based protocol enforcement tests whether a conclusion can be derived from extracted patient facts under explicit rules. Ontology-grounded semantic verification tests claims for consistency with, or entailment from, a formal biomedical ontology. Knowledge-graph-constrained reasoning tests whether individual claims are recorded in, or connected through, a structured entity-relation graph. None of the three checks whether the extracted facts are themselves correct. A closing subsubsection covers symbolic rule alignment, which builds these constraints into training.

\subsubsection*{Logic-Based Protocol Enforcement}
\label{subsubsec:logic_enforcement}

The most transparent version of this approach encodes protocols as formal logic programs and restricts LLM output to conclusions that can be formally derived from extracted patient facts. The LLM extracts information and handles language, and the symbolic engine has to derive every diagnostic or therapeutic claim. The derivation is only as reliable as the extraction, because the engine treats the facts it receives as true.

Tan et al.~\cite{Tan2026ProCDS} build this into ProCDS: the LLM extracts diagnostic rules from clinical guidelines and facts from the patient record, translates both into Prolog, and a backward-chaining engine derives the result. The detail that matters is the loop it forces. If the Prolog engine cannot produce a valid result, because extraction was incomplete, ambiguous, or inconsistent, the LLM has to revise the rules and facts before the engine runs again, so nothing is finalized without symbolic validation. On real records from 1,797 patients, classifying Obstructive Sleep Apnea Hypopnea Syndrome risk with 14 expert-validated rules, ProCDS reached 99.5\% accuracy on complete data and 85--95\% when 10\% of records had information masked, and its correction loop removed between a quarter and nearly all of the remaining errors depending on the condition~\cite{Tan2026ProCDS}. The engine cannot derive a conclusion that its rules do not support, but those rules are themselves generated by the LLM, so errors can still enter through fact extraction, rule encoding, or gaps in the rules' coverage; the symbolic layer constrains how conclusions are derived but does not remove these sources of error. Although the authors do not report it, the likely cost is latency, since incomplete data triggers repeated LLM-Prolog-LLM cycles. Kim and Wang~\cite{KimWang2025MentalHealthCLP} build a structurally similar system for psychiatric diagnosis using Constraint Logic Programming. An LLM translates diagnostic criteria from the ICD-11 Clinical Descriptions and Diagnostic Requirements (CDDR) into a logic program that domain experts can inspect and edit before it ever runs. The CLP engine then queries patient data against those rules to derive diagnoses that strictly follow the encoded criteria given the patient data supplied, a guarantee no standalone LLM can offer, though it does not extend to errors in how that data was recorded or extracted. Because the inspection happens at the specification layer, specification validation and execution stay separate. That adds a governance property purely neural architectures lack~\cite{KimWang2025MentalHealthCLP}.

The limiting case is deterministic constraint encoding, where the decision logic is fixed in advance and the LLM serves mainly as a language interface. Arriola-Montenegro et al.~\cite{ArriolaMontenegro2025AnemiaCare} approximate this in AnemiaCare HD, a hemodialysis anemia management assistant built as a GPT-based tool whose structured prompt, run at zero temperature with a fixed output template, constrains it to apply ESA kinetics rules, iron dosing thresholds, and timing safeguards. Across 600 simulated scenarios drawn from a single institutional protocol, an initial version with a loosely structured natural-language prompt matched the protocol exactly in 32\% of cases (96/300), and the redesigned deterministic prompt did so in 100\% (300/300)~\cite{ArriolaMontenegro2025AnemiaCare}. Because the rules are followed by an instructed LLM, not executed by a rule engine, this is an empirical result on simulated cases and not a formal guarantee; the authors note that a classical rule engine could implement the same logic. The gap between the two versions is the more informative number, since it shows how much protocol adherence depended on how tightly the prompt encoded the rules.

\subsubsection*{Ontology-Grounded Semantic Verification}
\label{subsubsec:ontology_verification}
 
Ontology-grounded verification works differently from logic-based enforcement. Rather than executing explicit rules, it checks semantic constraint satisfaction. It rejects any claim that contradicts the ontology's axioms. This is a consistency test, which is weaker than restricting output to what the ontology encodes, because a claim the ontology neither states nor contradicts passes. Requiring support means testing entailment instead (Section~\ref{sec:intro}). The paradigm suits domains where clinical knowledge is too relationally complex for standard Prolog rules and has to be formalized as OWL ontologies with description logic semantics.

The largest reported effect of ontological grounding in the reviewed literature comes from a GraphRAG framework for clinical question answering built on an ontology-grounded knowledge graph~\cite{OntologyCOVID2026}. Because the ontology serves as a source of grounding evidence for the LLM, not as a reasoner that tests individual claims, we classify the framework as evidence-grounded (Table~\ref{tab:verification_regimes}). A domain-specific RDF/OWL ontology and knowledge graph, built from clinical and hospital data from several Egyptian institutions, is integrated with the LLM to enforce structured semantic grounding during question answering. On 60 clinical questions scored against reference answers from five peer-reviewed hospital studies, the framework reached 98\% accuracy (59/60), against 37\% for ChatGPT-4 and 52\% for DeepSeek-R1, and its hallucination rate was 1.7\%, against approximately 63\% and 48\%~\cite{OntologyCOVID2026}. The authors attribute the improvement to embedding structured clinical semantics in the reasoning process. The evaluation set is small and drawn from a single national context, and whether gains of this size carry over to other structured clinical domains has not been tested.

Magana Vsevolodovna and Monti~\cite{OWLHermiT2025} push description logic reasoning further by wiring HermiT and Pellet reasoners into an LLM feedback loop. Candidate responses are checked for satisfiability against OWL axioms, and any violation triggers a refinement cycle that steers the LLM back toward semantic compliance. The foundation needs a caveat, though. Reasoning in expressive description logics has high worst-case complexity: concept satisfiability is NExpTime-complete for SHOIQ~\cite{Tobies2000Complexity}, the logic family underlying OWL 1 DL, and N2ExpTime-complete for SROIQ~\cite{Kazakov2008SROIQ}, the logic underlying OWL 2 DL. These are worst-case bounds, and optimized reasoners often perform well on real ontologies, but reasoning time on large, expressive ontologies is hard to predict. Full OWL 2 DL reasoning therefore suits offline audits or infrequent, high-stakes decisions better than live step-level supervision. Tractable profiles such as OWL 2 EL or OWL 2 RL trade expressivity for polynomial-time reasoning, and still need explicit justification against the encoded clinical structures. Whichever profile is used, a rejection is sound with respect to the encoded axioms: a claim flagged as inconsistent does contradict them. Acceptance means only that no contradiction was found. Neural self-consistency methods provide neither guarantee~\cite{OWLHermiT2025}.

\subsubsection*{Knowledge-Graph-Constrained Reasoning}
\label{subsubsec:kg_constrained}

The knowledge graph variant of ontological grounding encodes formal knowledge as a structured network of biomedical entities and typed relations, which allows both semantic constraint enforcement and fine-grained localization of hallucinations. Ontologies are checked at the axiom level and KGs at the claim level: each atomic assertion in the LLM's output is matched against the graph, and anything the graph does not support is vetoed. Two cautions apply. A direct edge records a stated relation, but a multi-hop path shows only that entities are connected; it supports a conclusion only if the relations along it compose to the claimed relation. And because clinical KGs are incomplete, a missing path shows that a claim is unsupported, not that it is false.

Sansford et al.~\cite{Sansford2024GraphEval} give a fine-grained version of claim-level checking in GraphEval, a general-domain hallucination detector. It converts LLM output into a graph of atomic claim triples (subject-relation-object) and uses a Natural Language Inference model to test each triple against the context the LLM was given. The result is more useful than a binary correctness judgment: it identifies which triples are unsupported, so a clinician or an automated module could fix the specific step that broke without discarding the rest of the reasoning. Its companion module, GraphCorrect, repairs most of the flagged triples through targeted replacement, which suggests that fine-grained localization can be more useful than rejecting a whole response~\cite{Sansford2024GraphEval}. GraphEval checks claims against the supplied context and was not evaluated on clinical data; using it against a curated clinical KG, as proposed in Section~\ref{subsec:integration}, would be an adaptation. Pusch and Conrad~\cite{Pusch2024KGLLM} move KG verification up to the query generation layer. They translate natural language into Cypher queries against a biomedical KG and use a symbolic algorithm to check syntax and schema alignment, so an incoherent query is corrected or rejected before it runs. Wang et al.~\cite{Wang2025MedKGI} apply strict ontological grounding to multi-turn diagnostic dialogue in MedKGI. It confines differential diagnosis generation to validated disease-symptom subgraphs and pairs this with an information-gain inquiry strategy and an OSCE state tracker to keep the evidence history coherent across turns. Anchoring hypotheses to paths in these validated subgraphs improved diagnostic dialogue efficiency by 30\% over baseline~\cite{Wang2025MedKGI}.

Extending KG-grounded reasoning to multi-hop inference runs into a problem: the candidate space grows steeply with depth. Cheng et al.~\cite{Cheng2026LogosKG} make high-hop traversal tractable with LogosKG, a hardware-optimized $k$-hop retrieval framework, and use it to show that the useful depth depends on the graph's topology. In zero-shot diagnosis, where the KG first filters unsupported LLM diagnoses and then supplies additional candidates, deeper retrieval generally improved performance on one dataset and changed it only moderately on another. When retrieved entities were instead added to the training context for supervised fine-tuning, cumulative multi-hop retrieval flooded the model with candidates (on average over 200 diagnosis entities within two hops and over 42,000 within five) and pushed performance below standard fine-tuning. The authors conclude that simple fine-tuning over deep KG structures is not enough and point to iterative, multi-agent frameworks as the direction to pursue. In KG-TRACES, Wu et al.~\cite{Wu2025KGTRACES} drop the post-hoc veto layer entirely and use attribution-aware multi-task fine-tuning that trains the LLM to predict symbolic relation paths and generate step-level reasoning, tagging each step with its origin, the KG or internal inference.

Hu et al.~\cite{Hu2025GraphRAG} make the same shift toward iterative control and build it into a hepatology decision support system. Their Agentic Graph RAG pairs a clinically verified KG with a deterministic self-correction loop. It translates queries to Cypher and validates them semantically before execution, checks whether the retrieved context is sufficient through dual-model self-consistency voting, and falls back to a non-LLM vector-similarity adjudicator whenever the models disagree, so the decision to stop retrieving rests on an external criterion instead of either model's own judgment, although that criterion can itself be wrong. When the context is insufficient, a deterministic strategy optimizer forces another retrieval pass. On a set of 30 custom hepatology questions, this model-independent governance reached 0.94 faithfulness and 0.92 context recall, above both ungrounded LLMs and standard RAG baselines~\cite{Hu2025GraphRAG}.

\subsubsection*{Symbolic Rule Alignment and Preference Learning}
\label{subsubsec:rule_alignment}

A complementary paradigm skips runtime verification and builds clinical rule compliance directly into the training objective. It aligns the model's parametric behavior toward rule adherence through preference learning, fine-tuning on rule-derived datasets, or by encoding formal logic into the training signal itself, with no symbolic veto layer at inference.

Wang et al.~\cite{RuleAlign2024} build this in RuleAlign, which fine-tunes an LLM through preference learning on rule-bounded physician-patient dialogues, with a training objective that directly penalizes any output that violates an encoded diagnostic rule. Rule adherence is folded into the parametric weights themselves. The runtime symbolic engine is skipped entirely, and latency falls well below what iterative verification architectures need. Preference-learned alignment is still probabilistic: the model follows rules with high probability, not certainty, so it cannot offer the guarantee a deterministic symbolic checker gives, that every output is consistent with the encoded rules. The choice between alignment-time and inference-time enforcement is therefore one of computational efficiency against formal verifiability~\cite{RuleAlign2024}. For high-stakes decisions such as drug dosing, surgical eligibility, and crisis intervention, we argue that runtime enforcement of the encoded rules is needed. Alignment-time compliance may be enough for lower-stakes advisory functions.
 
\subsection{Dynamic Logic Verification}
\label{subsec:dynamic_verification}

The approaches in Section~\ref{subsec:formal_logic} are static by design. They check LLM output against a fixed rule base, ontology, or knowledge graph that encodes clinical knowledge as of a specific point in time. Dynamic logic verification enforces logical entailment across individual inferential steps in real time instead. Theorem provers, abductive logic programming, and automated formalization check whether the formalized reasoning path is itself valid, not only whether its conclusion is consistent with a knowledge base. The guarantee applies to the formalization: if the translation from natural language into logic is wrong, a valid proof certifies the wrong statement.

\subsubsection*{Theorem Proving and Autoformalization}
\label{subsubsec:theorem_proving}

Quan et al.~\cite{Quan2024TheoremProving} built Explanation-Refiner, a general-domain system, to meet the strictest standard for coupling LLMs with automated theorem provers: a mechanically verified proof that a formalization of an LLM's output is entailed by its premises. It converts LLM-generated natural language into first-order logic through Neo-Davidsonian event semantics and hands the result to Isabelle/HOL for verification. A failed proof attempt triggers a loop. It may reflect a logical error in the explanation, but also missing premises, a faulty translation into logic, or the prover's search or time limits, so it shows that the explanation could not be verified, not that it is invalid. The prover sends the specific failed step back to the LLM for targeted refinement, and the cycle repeats until the proof goes through or a maximum iteration count is reached. On the e-SNLI benchmark with GPT-4, logically valid explanations rose from a 36\% baseline to 84\% after iterative refinement~\cite{Quan2024TheoremProving}. The 36\% baseline indicates how often fluent LLM explanations could not be verified: nearly two-thirds of those tested could not be proved, for any of the reasons above. The 84\% that refinement reaches rests on mechanical proof of the formalized explanation, a stronger form of evidence than the audit-pattern coverage of evidence-traceability systems, which can fail silently once reasoning drifts out of distribution. The two paradigms complement each other. Full theorem proving governs systems that can be formally specified well enough to support it, and evidence-traceability checking scales to the heterogeneous knowledge bases that first-principles formalization cannot reach.

An et al.~\cite{AutoformalizationNS2025} apply autoformalization to natural-language policy verification outside the clinical domain, reaching over 99\% soundness in logical validity detection. Their architecture runs redundant, parallel formalization passes and cross-checks them for semantic equivalence, an explicit trade of computational cost for stronger formal verification. Healthcare appears in that work only as an example domain, so its clinical value is untested.

\subsubsection*{Abductive Logic and Counter-Abduction}
\label{subsubsec:abductive_logic}

Abductive Logic Programming (ALP) offers a methodologically distinct route to dynamic verification, one that tests whether competing explanations fit the evidence as well as the proposed one, in addition to whether that explanation is logically consistent. It treats LLM-generated clinical inferences as abductive hypotheses, meaning candidate explanations for observed patient states. Each one then faces counter-abduction: competing hypotheses that could equally explain the evidence are generated systematically and compared under the preference criteria the procedure implements.

The approach of Galitsky and Rybalov~\cite{ALPVerification2026}, originally developed for industrial process control, uses Prolog meta-interpreters to generate counter-hypotheses that actively defeat inconsistent LLM reasoning. Formally, abduction works over a background theory $T$, a set of abducibles $A$ (candidate hypotheses), and an observation $O$. The mechanism finds a minimal $\Delta \subseteq A$ such that $T \cup \Delta \models O$, while keeping $T \cup \Delta$ consistent. Moving this architecture into clinical deployment brings two limitations. First, clinical knowledge bases suffer from partial observability (e.g., undocumented comorbidities or missing laboratory values), so any explanation found holds relative to the available evidence, not to the complete patient state. Second, meta-interpreter search over unrestricted first-order Horn theories is computationally intractable, so implementations need depth bounds that explicitly sacrifice completeness for termination. The advantage over deductive checking survives these limits: a claim can follow logically from the theory and still be a poor explanation if a competing hypothesis also explains the observations. The abductive condition above establishes only that $\Delta$ explains $O$ relative to $T$ while remaining consistent with it. It says nothing about real-world causal truth, and preferring one explanation over another requires an explicit ordering over hypotheses, such as minimality or a cost function. Within those bounds, counter-abduction can reject an LLM explanation when the implemented ordering prefers a competing one, although the Prolog meta-interpreter becomes a scalability bottleneck during complex multi-hypothesis evaluation~\cite{ALPVerification2026}.

Haibu MMIA~\cite{Zhang2025Haibu} offers a lighter, LLM-based alternative for healthcare administration tasks. It breaks tasks into atomic steps, has an independent LLM instance audit each step for logical coherence and evidence traceability, and stores audited reasoning chains as reusable ``theorems'' retrieved by vector similarity; no proof assistant or formal logic checker is involved. The authors report that reuse substantially lowers later token costs (their headline figure is about 85\%, although the cost figures in the paper imply a smaller saving) and that the audit caught over 98\% of errors with few false positives across four healthcare administration domains~\cite{Zhang2025Haibu}. These are the authors' own evaluations, and the checks are not formal proofs. The two approaches nonetheless suggest a design principle for clinical verification architectures. Full ALP verification suits high-stakes, low-frequency decisions (e.g., surgical eligibility, crisis intervention), where an explanation should be checked against explicit alternatives. LLM-audited decomposition with reuse may serve higher-volume workflows, where efficiency matters more and formal guarantees are not required.

\subsubsection*{Multimodal and Biomarker-Linked Verification}
\label{subsubsec:multimodal_ns}

Extending the neuro-symbolic verification paradigm to multimodal clinical domains (radiology, pathology, ophthalmology) requires that Vision-Language Model (VLM) outputs be grounded in both formal knowledge structures and verified visual evidence. This speaks directly to the multimodal hallucination failure modes documented in Section~\ref{sec:core_problem}, which come from the decoupling of visual feature encoders and textual claim generation.

Wang and Qin~\cite{Wang2026Explainable} apply multimodal grounding within a neuro-symbolic framework for corneal topography interpretation and refractive surgery screening. They do not rely on a vision-language model for end-to-end diagnostic reasoning. A hybrid CNN-Vision Transformer module extracts spatial features and aligns them directly into a symbolic corneal knowledge graph, and diagnostic classification then runs through Bayesian probabilistic inference across these formal nodes (linking biometric parameters, ectatic disease states, and surgical eligibility criteria). The neural components are limited to visual feature extraction and natural language generation, and authority over clinical conclusions rests entirely with the symbolic reasoning layer. In a pilot evaluation on 20 eyes, the framework achieved an area under the curve of 0.95 for early keratoconus detection and a 0.90 F1 score for surgical eligibility classification. The result is promising for knowledge-constrained inference over multimodal inputs, but the sample is too small to establish it~\cite{Wang2026Explainable}.

Hong et al.~\cite{Hong2026RECAP} address pathology diagnostics through RECAP-PATH, which moves multimodal large language models from passive pattern recognition to active reasoning. The framework uses a two-phase automated prompt optimization loop with expert feedback rather than a strict symbolic veto. That forces the model to state explicit morphological evidence before classification, which ties the final diagnostic prediction to verified textual reasoning and produces audit-ready clinical reports.

Lopukhova et al.~\cite{Lopukhova2025CDSS} apply neuro-symbolic separation to ophthalmic imaging with a two-stage clinical decision support system for age-related macular degeneration staging. A deep neural network first extracts specific imaging biomarkers and produces temperature-calibrated confidence scores. A fuzzy logic inference module then combines these scores with expert-defined diagnostic rules. Stage probabilities are constrained to expert-defined biomarker patterns, and this decoupled architecture achieved a 63.3\% relative accuracy increase over direct neural classification while remaining interpretable~\cite{Lopukhova2025CDSS}.

\subsubsection*{The Safety--Efficiency Trade-off and the Limits of Neuro-Symbolic Grounding}
\label{subsubsec:ns_tradeoffs}
 
The neuro-symbolic architectures reviewed in this section share a trade-off: the computational overhead they add grows with the depth of their formal verification. Prolog backward-chaining iterations, description logic satisfiability testing in expressive OWL ontologies, multi-hop knowledge graph path validation, and automated theorem proving all add sequential processing stages, so latency rises compared with single-pass neural generation. The reported benefits within formally encoded domains suggest that this cost can be worthwhile for high-stakes deployment. They include 99.5\% risk-classification accuracy on complete OSAHS data~\cite{Tan2026ProCDS} and a hallucination rate reduced by over 61 percentage points in ontology-grounded clinical question answering~\cite{OntologyCOVID2026}; MedKGI's 30\% gain~\cite{Wang2025MedKGI} concerns dialogue efficiency, not safety. These are single-study results, mostly from simulated or small-sample settings, and they are not directly comparable with neural baselines evaluated elsewhere. They indicate what domain-bounded symbolic constraints can achieve, but they do not yet establish a general safety advantage over purely neural architectures.

A secondary limitation is domain specificity. Every reviewed neuro-symbolic system remains constrained by its foundational knowledge representation: a Prolog rule base for OSAHS, an RDF/OWL clinical ontology built from Egyptian hospital data, a corneal knowledge graph for refractive surgery. Building and maintaining that symbolic layer takes formal knowledge engineering and continuous updating to follow evolving clinical guidelines, which mirrors the adaptation challenge discussed in Section~\ref{sec:prm}. Kiruluta~\cite{Kiruluta2025DecisionTree} proposes one way to reduce this bottleneck: trainable decision-tree and random-forest oracles embedded in a symbolic reasoning layer and coordinated with LLM planning by a central orchestrator, so that protocol-like rules can be learned instead of hand-crafted. The reported gains, 7.2\% in entailment consistency and 5.3\% in multistep reasoning accuracy, come from general reasoning benchmarks (ProofWriter, GSM8K, and ARC), and clinical decision support is discussed only as an application. Learned symbolic oracles therefore remain a promising but untested route to scaling clinical verification without exhaustive manual formalization~\cite{Kiruluta2025DecisionTree}. Even so, generalizing neuro-symbolic grounding beyond specific training domains remains unresolved. The field has yet to demonstrate an architecture that keeps formal verification guarantees across the unbounded scope of a general-purpose clinical assistant.

Bhattacharya and Ashish~\cite{NeurosymbolicSafety2025} propose a dual-process framework for neuro-symbolic AI. The neural network is a fast, intuitive System 1 reasoner and the symbolic engine a deliberative System 2 verifier. That presents the computational latency trade-off as a coherent cognitive architecture, in which advanced systems should activate symbolic verification selectively, based on confidence estimates or clinical risk stratification. Kolli~\cite{FairnessNS2025} applies this principle in a hybrid ethical failsafe architecture that uses symbolic fairness-aware rule checking to halt clinical plans that violate predefined risk thresholds or compliance guardrails. Establishing principled activation criteria for this symbolic layer (when to invoke full description logic reasoning, lightweight knowledge graph verification, or pure neural generation) remains the defining open problem for making neuro-symbolic grounding practical across general clinical deployment.

The literature has not addressed a third deployment constraint, data governance. The symbolic mechanisms reviewed here, particularly ontology-based query checking and knowledge graph constraints, typically require sending patient-specific claims to a centralized verification layer. In multi-institutional deployments this can conflict with HIPAA and the GDPR, which restrict the disclosure and cross-border transfer of health data to specific legal bases and safeguards. A neuro-symbolic architecture that gives strong safety guarantees within a single institution may therefore be non-compliant at scale. Privacy-preserving approaches, including federated symbolic reasoning and differential privacy mechanisms adapted for ontology checking, are an open research direction. Until federated-compatible neuro-symbolic architectures emerge, the safety guarantees documented throughout this section remain limited to single-institution settings or to data-sharing arrangements with a valid legal basis.

\subsection{Integration Architecture: Mapping Agentic Failure Modes to Symbolic Intervention Points}
\label{subsec:integration}

\begin{table}[H]
\scriptsize
\setlength{\tabcolsep}{4pt}
\renewcommand{\arraystretch}{1.0}
\caption{The five verification regimes defined in Section~\ref{sec:intro}, together with abductive verification, with their representative systems, guarantee strength, and bounding factor across Sections~\ref{sec:prm} and~\ref{sec:neurosymbolic}. Deductive systems are split into runtime checkers and training-time rule rewards. Abductive verification is listed in its own row because it tests whether hypothesized premises explain the observations, not whether a conclusion follows from the evidence. Conceptual frameworks are listed separately because they report no evaluation. Superscripts show the source type: P = peer-reviewed, PP = preprint, CF = conceptual or speculative framework (following the source classification defined in Section~\ref{sec:methods}); \dag\ = evaluated outside the clinical domain.}
\label{tab:verification_regimes}
\begin{tabularx}{\linewidth}{p{0.10\linewidth} X p{0.13\linewidth} p{0.17\linewidth} >{\centering\arraybackslash}p{0.028\linewidth}}
\toprule
\textbf{Regime} &
\textbf{Representative Systems} &
\textbf{Guarantee Strength} &
\textbf{Bounding Factor} &
\textbf{Sec.} \\
\midrule

\textbf{Deductive (runtime)} &
ProCDS\textsuperscript{P}~\cite{Tan2026ProCDS}, CLP psychiatric CDSS\textsuperscript{PP}~\cite{KimWang2025MentalHealthCLP}, Explanation-Refiner\textsuperscript{P\,\dag}~\cite{Quan2024TheoremProving}, autoformalization policy checker\textsuperscript{PP\,\dag}~\cite{AutoformalizationNS2025} &
Conclusion entailed by the encoded rules and the supplied facts &
Correctness of the encoded rules (including any the LLM generated), of the extracted facts, and of any formalization &
\ref{sec:neurosymbolic} \\

\midrule

\textbf{Abductive (runtime)} &
ALP counter-abduction\textsuperscript{P\,\dag}~\cite{ALPVerification2026} &
Explanatory adequacy and consistency, conditional on the hypothesized premises and the implemented preference ordering &
Background theory, set of abducibles, preference ordering, and search depth bounds &
\ref{sec:neurosymbolic} \\

\midrule

\textbf{Deductive (training-time rewards)} &
VPRM\textsuperscript{PP}~\cite{Pronesti2026VPRM}, Med-U1\textsuperscript{PP}~\cite{Zhang2025MedU1} &
Rule-checked rewards during training; no runtime guarantee &
Coverage of the encoded rules; residual reward hacking &
\ref{sec:prm} \\

\midrule

\textbf{Ontology-based} &
HermiT/Pellet DL reasoner\textsuperscript{PP}~\cite{OWLHermiT2025} &
Rejections sound under OWL/DL semantics; acceptance means consistent, not entailed, unless entailment is tested &
How complete and up to date the ontology is; computing cost grows with expressivity &
\ref{sec:neurosymbolic} \\

\midrule

\textbf{Evidence-grounded} &
Med-PRM\textsuperscript{P}~\cite{Yun2025MedPRM}, ontology-grounded GraphRAG framework\textsuperscript{P}~\cite{OntologyCOVID2026}, GraphEval\textsuperscript{P\,\dag}~\cite{Sansford2024GraphEval}, Pusch and Conrad's Cypher checker\textsuperscript{P}~\cite{Pusch2024KGLLM}, MedKGI\textsuperscript{PP}~\cite{Wang2025MedKGI}, LogosKG\textsuperscript{PP}~\cite{Cheng2026LogosKG}, KG-TRACES\textsuperscript{PP}~\cite{Wu2025KGTRACES}, Agentic Graph RAG\textsuperscript{P}~\cite{Hu2025GraphRAG}, Haibu MMIA\textsuperscript{PP}~\cite{Zhang2025Haibu}, corneal knowledge graph\textsuperscript{P}~\cite{Wang2026Explainable}, MedReason\textsuperscript{PP}~\cite{Wu2025MedReason}, Fleming-R1\textsuperscript{PP}~\cite{FlemingR12025}, process reward agents\textsuperscript{PP}~\cite{Sohn2026PRA} &
Auditable and updatable, but no formal entailment &
How well the corpus or graph covers the topic, and retrieval quality &
\ref{sec:prm}, \ref{sec:neurosymbolic} \\

\midrule

\textbf{Probabilistic learned} &
Clinical note PRM\textsuperscript{P}~\cite{Wang2024ClinicalNotePRM}, radiology report PRM\textsuperscript{PP}~\cite{Thomas2025RadiologyPRM}, HuatuoGPT-o1\textsuperscript{PP}~\cite{Chen2024HuatuoGPT}, MedS$^3$\textsuperscript{P}~\cite{Jiang2025MedS3}, MiraMind/Mindora\textsuperscript{PP}~\cite{Xiao2025MentraSuite}, bilateral RL psychological reasoning\textsuperscript{P}~\cite{BilatRL2026}, AMD fuzzy staging\textsuperscript{P}~\cite{Lopukhova2025CDSS}, decision tree/random forest oracles\textsuperscript{PP\,\dag}~\cite{Kiruluta2025DecisionTree} &
Statistical, and depends on the data distribution &
Coverage and quality of training or reward-annotation data &
\ref{sec:prm}, \ref{sec:neurosymbolic} \\

\midrule

\textbf{Structural process-level} &
MedCEG\textsuperscript{PP}~\cite{Mu2025MedCEG}, MedVLM-R1\textsuperscript{PP}~\cite{Pan2025MedVLMR1}, MedCCO\textsuperscript{PP}~\cite{Rui2025MedCCO}, P2S\textsuperscript{PP\,\dag}~\cite{Zhong2026P2S}, Psyche-R1\textsuperscript{PP}~\cite{Dai2025PsycheR1}, RuleAlign\textsuperscript{PP}~\cite{RuleAlign2024}, RECAP-PATH\textsuperscript{P}~\cite{Hong2026RECAP}, AnemiaCare HD (prompt-constrained)\textsuperscript{P}~\cite{ArriolaMontenegro2025AnemiaCare} &
A learned or instructed tendency, with no external judge &
Only limits from the training format or prompt; no runtime check against a specification &
\ref{sec:prm}, \ref{sec:neurosymbolic} \\

\midrule

\textbf{Conceptual frameworks} &
Dual-process S1/S2 framework\textsuperscript{CF}~\cite{NeurosymbolicSafety2025}, fairness failsafe\textsuperscript{CF}~\cite{FairnessNS2025} &
None demonstrated &
No reported evaluation &
\ref{sec:neurosymbolic} \\

\bottomrule
\end{tabularx}
\renewcommand{\arraystretch}{1}
\end{table}
\vspace{-0.3cm}

The paradigms reviewed in Sections~\ref{subsec:formal_logic} and~\ref{subsec:dynamic_verification} mostly work as standalone verification architectures. They become clinically useful only when deployed as targeted constraint layers at the structural boundaries of the agentic pipelines described in Section~\ref{sec:agentic}. This subsection maps these symbolic interventions to the three primary failure categories documented earlier.

Sequential tool-augmented architectures such as EHRAgent~\cite{Shi2024EHRAgent} are exposed to two vulnerabilities during multi-step reasoning. At the fact-assertion boundary, unverified patient extractions risk poisoning later inferential steps. The forced-iteration design of ProCDS~\cite{Tan2026ProCDS} (Section~\ref{subsubsec:logic_enforcement}) partly addresses this by refusing to commit any conclusion that a Prolog engine cannot derive from the extracted facts and protocol rules, and by forcing the LLM to revise them when derivation fails. It does not detect an extracted value that is wrong but plausible, so a full fix also needs a provenance check of each fact against the source record. At the query-execution boundary, code-generated database queries risk executing hallucinated structural relationships. The symbolic validation framework of Pusch and Conrad~\cite{Pusch2024KGLLM} (Section~\ref{subsubsec:kg_constrained}) reduces this risk by checking LLM-generated Cypher queries against the target knowledge graph schema before execution. A schema-valid query can still ask the wrong question, so this check prevents malformed queries, not incorrect ones. Together these layers would narrow the main cascade pathways of ungrounded sequential agentic pipelines.

In multi-agent collaborative architectures, Zhu et al.~\cite{MedAgentAudit2025} show that false consensus often overrides independent clinical verification (Section~\ref{subsubsec:multi_agent}). Their MedAgentAudit framework finds that collaborative synthesis suppresses correct minority views through authority bias and unverified claim repetition. Authority bias alone climbs from 35.30\% at the first synthesis step to 68.75\% by the third. Countering this requires symbolic intervention at the inter-agent communication boundary. We propose adapting the triple-level localization of GraphEval~\cite{Sansford2024GraphEval} (Section~\ref{subsubsec:kg_constrained}) for this purpose; GraphEval itself checks triples against the supplied context, and its use with a clinical knowledge graph has not been tested. Before any agent's reasoning reaches the shared debate context, the output would be broken into atomic (subject, relation, object) triples, and only assertions that the clinical knowledge graph records, or implies under defined relation semantics, would enter the consensus pool. Filtering claims independently of any agent's confidence would limit how far conformity can amplify unsupported claims, and would give a correct minority view a basis that does not depend on how many agents hold it. It would not remove authority bias, which can still act through supported but irrelevant claims, omissions, or misleading synthesis. An incomplete KG could also reject a correct minority claim, so rejected claims should be escalated for review, not silently deleted.

In specialized orchestration frameworks such as MMedAgent~\cite{Li2024MMedAgent} and AgentClinic~\cite{Schmidgall2024AgentClinic} (Section~\ref{subsubsec:orchestration}), an erroneous planning decision early in a multi-modal pipeline can corrupt all downstream inference. That calls for symbolic intervention at the pipeline composition boundary, where the reasoning LLM chooses which downstream tool or specialist agent to invoke. The orchestrator design of Kiruluta~\cite{Kiruluta2025DecisionTree} (Section~\ref{subsubsec:ns_tradeoffs}), in which trained decision-tree and random-forest oracles check an LLM planner's hypotheses before execution, suggests one way to target this boundary. In the proposed framework, such oracles would intercept planning outputs before any external tool is invoked and veto tool-use decisions that violate encoded clinical protocols. Misrouted tool selections that those protocols cover, and that the oracles correctly detect, would then be stopped before they cascade downstream; others would pass. This use is a proposal of this review, since the oracles have so far been evaluated only on general reasoning benchmarks.

The three mappings define this review's framework. A cascade that spreads through fact assertion needs a provenance check and a derivation check at the extraction boundary. One driven by inter-agent consensus needs formal intervention at the communication boundary. One that starts from erroneous tool selection needs logical constraints at the planning boundary. The symbolic architectures of Sections~\ref{subsec:formal_logic} and~\ref{subsec:dynamic_verification} therefore act as targeted interventions at the architectural vulnerabilities behind the failure modes of Section~\ref{sec:agentic}.

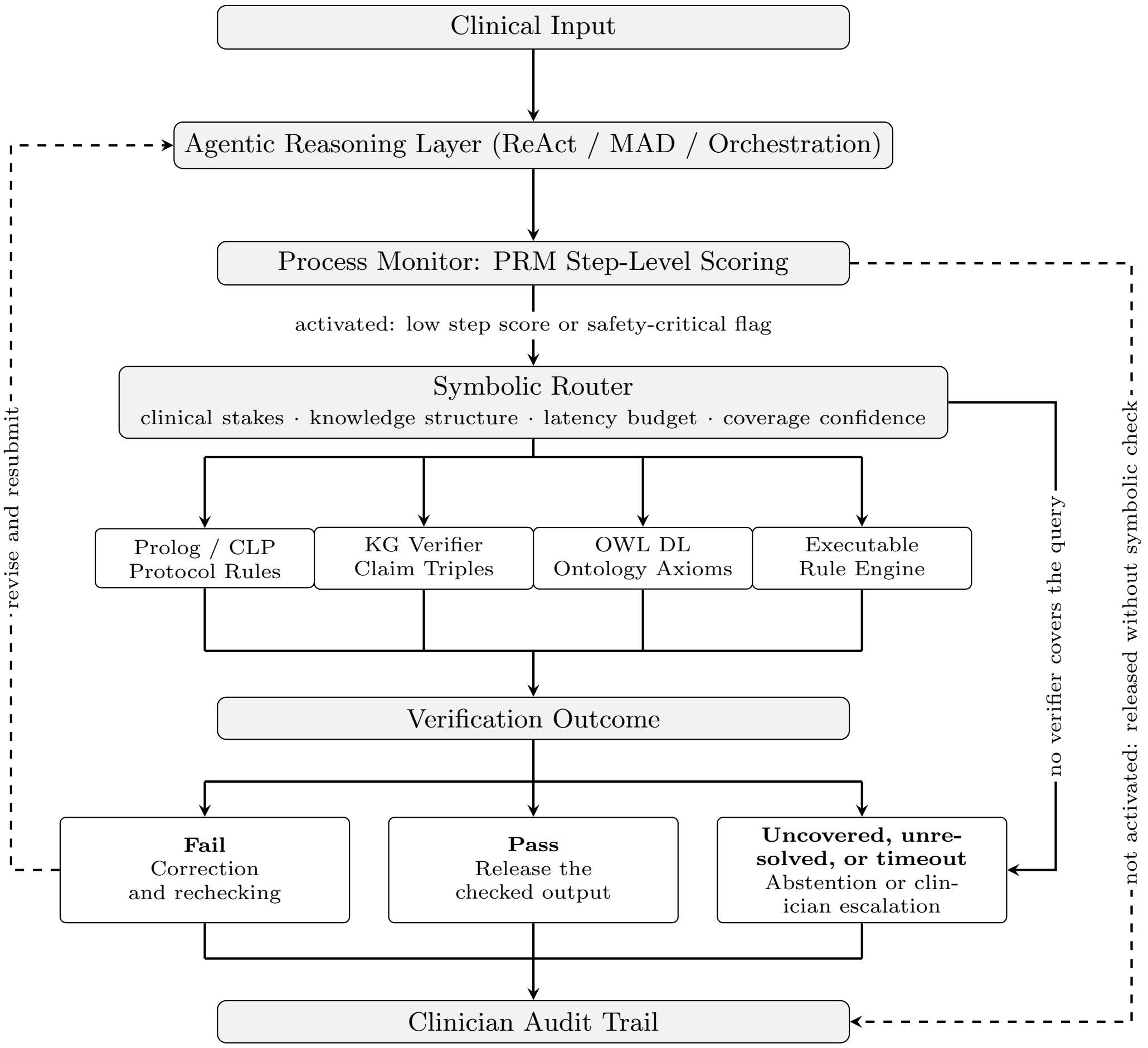
\begin{figure}[H]
\centering
\begin{tikzpicture}[
  box/.style={rectangle, draw, rounded corners=3pt,
    minimum width=7.5cm, minimum height=0.5cm,
    text centered, fill=gray!10, font=\small},
  vbox/.style={rectangle, draw, rounded corners=2pt,
    minimum width=2.6cm, minimum height=0.7cm,
    text centered, fill=white, font=\scriptsize, align=center},
  obox/.style={rectangle, draw, rounded corners=2pt,
    text width=3.2cm, minimum height=1.25cm,
    fill=white, font=\scriptsize, align=center},
  lab/.style={font=\scriptsize, fill=white, inner sep=1.5pt, align=center},
  arr/.style={-stealth, thick},
  darr/.style={-stealth, thick, dashed}
]
\node[box] (inp)    at (0,  0.0) {Clinical Input};
\node[box] (agent)  at (0, -1.4) {Agentic Reasoning Layer (ReAct / MAD / Orchestration)};
\node[box] (prm)    at (0, -2.8) {Process Monitor: PRM Step-Level Scoring};
\node[box, text width=9.6cm, align=center, minimum height=0.85cm] (router) at (0, -4.45) {Symbolic Router\\[-1pt]{\scriptsize clinical stakes $\cdot$ knowledge structure $\cdot$ latency budget $\cdot$ coverage confidence}};
\node[vbox] (prolog) at (-3.9, -6.3) {Prolog / CLP\\Protocol Rules};
\node[vbox] (kg)     at (-1.3, -6.3) {KG Verifier\\Claim Triples};
\node[vbox] (owl)    at ( 1.3, -6.3) {OWL DL\\Ontology Axioms};
\node[vbox] (det)    at ( 3.9, -6.3) {Executable\\Rule Engine};
\node[box] (outc)  at (0, -8.2) {Verification Outcome};
\node[obox] (fail) at (-3.9, -10.0) {\textbf{Fail}\\Correction and rechecking};
\node[obox] (pass) at ( 0.0, -10.0) {\textbf{Pass}\\Release the checked output};
\node[obox] (abst) at ( 3.9, -10.0) {\textbf{Uncovered, unresolved, or timeout}\\Abstention or clinician escalation};
\node[box] (audit) at (0, -11.8) {Clinician Audit Trail};
%--- Main pipeline
\draw[arr] (inp) -- (agent);
\draw[arr] (agent) -- (prm);
\draw[arr] (prm) -- node[lab] {activated: low step score or safety-critical flag} (router);
%--- Router fan-out
\draw[thick] (router.south) -- (0,-5.1);
\draw[thick] (-3.9,-5.1) -- (3.9,-5.1);
\draw[arr] (-3.9,-5.1) -- (prolog.north);
\draw[arr] (-1.3,-5.1) -- (kg.north);
\draw[arr] ( 1.3,-5.1) -- (owl.north);
\draw[arr] ( 3.9,-5.1) -- (det.north);
%--- Verifier fan-in
\draw[thick] (prolog.south) -- (-3.9,-7.4);
\draw[thick] (kg.south)     -- (-1.3,-7.4);
\draw[thick] (owl.south)    -- ( 1.3,-7.4);
\draw[thick] (det.south)    -- ( 3.9,-7.4);
\draw[thick] (-3.9,-7.4) -- (3.9,-7.4);
\draw[arr] (0,-7.4) -- (outc.north);
%--- Outcome fan-out
\draw[thick] (outc.south) -- (0,-8.95);
\draw[thick] (-3.9,-8.95) -- (3.9,-8.95);
\draw[arr] (-3.9,-8.95) -- (fail.north);
\draw[arr] ( 0.0,-8.95) -- (pass.north);
\draw[arr] ( 3.9,-8.95) -- (abst.north);
%--- All outcomes are logged
\draw[thick] (fail.south) -- (-3.9,-11.05);
\draw[thick] (pass.south) -- ( 0.0,-11.05);
\draw[thick] (abst.south) -- ( 3.9,-11.05);
\draw[thick] (-3.9,-11.05) -- (3.9,-11.05);
\draw[arr] (0,-11.05) -- (audit.north);
%--- Correction feedback loop
\draw[darr] (fail.west) -- (-6.2,-10.0) -- node[lab, rotate=90] {revise and resubmit} (-6.2,-1.4) -- (agent.west);
%--- No verifier covers the query
\draw[arr] (router.east) -- (6.2,-4.45) -- node[lab, rotate=90] {no verifier covers the query} (6.2,-10.0) -- (abst.east);
%--- Symbolic check not activated
\draw[darr] (prm.east) -- (7.1,-2.8) -- node[lab, rotate=90] {not activated: released without symbolic check} (7.1,-11.8) -- (audit.east);
\end{tikzpicture}
\caption{Proposed framework for verifiable clinical reasoning. A clinical input passes through an agentic reasoning layer and a step-level PRM monitor. Symbolic checking is not applied to every query: it is activated when a reasoning step scores below a deployment-calibrated threshold or when the query is flagged as safety-critical, and outputs that do not meet either condition are released without a symbolic check and logged as unchecked. For activated queries, the router selects one of four verifier classes according to four criteria (clinical stakes, knowledge structure, latency budget, and coverage confidence); queries that no verifier covers go directly to abstention or clinician escalation. Verification has three outcomes. A claim that passes is released as a checked output. A claim that fails is returned to the agentic layer for correction and rechecking (dashed loop). A claim that stays unresolved, or whose check exceeds the latency budget, leads to abstention or escalation. Every outcome is logged to a clinician-accessible audit trail.}
\label{fig:unified_framework}
\end{figure}

Figure~\ref{fig:unified_framework} lays out the pipeline implicit in these boundary mappings. Four criteria govern the routing decision at the symbolic layer. \textit{Clinical stakes} asks whether an error is irreversible or safety-critical, and if so the query goes to the highest-guarantee verifier available. \textit{Knowledge structure} asks whether the domain is covered by formal protocol rules, an ontological hierarchy, a biomedical entity graph, or an enumerable rule set. \textit{Latency budget} asks whether the deployment context can tolerate the multi-pass iteration that OWL DL satisfiability or multi-hop KG path validation requires. \textit{Coverage confidence} asks whether the query falls within the trained or curated scope of any available verifier. Table~\ref{tab:routing_criteria} puts these criteria into practice. Symbolic checking is not applied to every query, and the framework separates five cases. \textit{Verification} is triggered when the PRM scores the current reasoning step below a deployment-calibrated threshold (Section~\ref{sec:prm}) or when the agentic layer flags the query as safety-critical; a safety-critical flag routes the query to the strongest applicable verifier or to clinician escalation and does not by itself count against the claim. Outputs that meet neither condition are released without a symbolic check and are logged as unchecked. \textit{Acceptance} occurs when the selected verifier's check succeeds; the output is released as checked, with the check recorded in the audit trail. \textit{Rejection} occurs when the selected verifier finds the claim contradicted by its formal knowledge base or, where support is required, not derivable from it or not recorded in it, either directly or under defined relation semantics. \textit{Correction} follows a rejection when the verifier localizes the failing fact or step and the agentic layer can revise it within the latency budget. \textit{Structured abstention} follows when no verifier covers the query, when correction fails, or when the budget runs out, so the symbolic boundary does not silently hand out-of-domain queries to unconstrained neural generation.

\begin{table}[H]
\footnotesize
\setlength{\tabcolsep}{5pt}
\renewcommand{\arraystretch}{1.2}
\caption{Symbolic verifier routing criteria for the proposed integration framework (Figure~\ref{fig:unified_framework}). All verifiers are proposed components. In particular, the executable rule engine is a proposal: the closest reviewed system, AnemiaCare HD~\cite{ArriolaMontenegro2025AnemiaCare}, applies a hemodialysis protocol through a constrained prompt, not through an executable rule engine.}
\label{tab:routing_criteria}
\begin{tabularx}{\linewidth}{p{0.24\linewidth} p{0.16\linewidth} p{0.17\linewidth} p{0.33\linewidth}}
\toprule
\textbf{Decision Domain} & \textbf{Verifier} & \textbf{Expected Latency} & \textbf{Primary Veto Condition} \\
\midrule
Formal protocol coverage (drug dosing, surgical eligibility) & Prolog / CLP & Medium (iterative re-extraction) & Conclusion not derivable from the protocol rules and extracted facts, or an extracted fact not traceable to the source record \\
\midrule
Structured ontological domain (rare disease phenotype, infectious disease semantics) & OWL DL & Potentially high; profile- and ontology-dependent & Candidate claim inconsistent with, or (where required) not entailed by, the ontology axioms \\
\midrule
Multi-hop relational claim (differential diagnosis, drug interaction) & KG Verifier & Medium & Claim triple neither recorded in the KG nor implied by composable relations, or entity absent from the graph \\
\midrule
Fully enumerable safety-critical rule set (e.g., a hemodialysis dosing protocol) & Executable rule engine & Expected to be minimal (not measured) & Any output that departs from the encoded rule set (checked on every query) \\
\midrule
Out-of-domain query (no formal coverage available) & Abstention & N/A & Default fallback for all unrouted queries \\
\bottomrule
\end{tabularx}
\end{table}

Four structural failure conditions limit the framework's operating range. \textit{Ontology staleness}: a Prolog rule base or OWL ontology that encodes superseded clinical guidelines produces formally valid but clinically incorrect verdicts, which calls for the continuous specification monitoring proposed in Section~\ref{subsec:ns_eval}. \textit{Entity extraction failure}: errors in the upstream neural extraction layer put corrupted facts into the formal reasoning base. A wrong but plausible value still yields a valid derivation and passes silently; only a missing or implausible one triggers the forced re-extraction loop, where repeated failures on ambiguous clinical text can use up the latency budget without resolving anything. \textit{Guideline conflict}: a patient whose care crosses domains governed by partly contradictory protocols can generate symbolic constraints that cannot all be satisfied at once, and resolving that takes explicit arbitration that no reviewed architecture provides. \textit{Latency exceedance}: OWL DL satisfiability checking under expressive profiles, and multi-hop KG path validation at high query volumes, can exceed the time constraints of emergency or intensive-care settings, so profile-restricted reasoning or pre-computed verification caches are needed as operational fallbacks.

One qualification applies to the framework as a whole. The architecture in Figure~\ref{fig:unified_framework} is a conceptual specification. The studies reviewed in Sections~\ref{sec:prm} and~\ref{sec:neurosymbolic} evaluated its individual components in isolation, and their integration under a shared routing architecture is a research objective that has not been validated clinically. Deployment would first need prospective end-to-end evaluation across diverse patient populations, regulatory approval under the frameworks identified in Section~\ref{sec:conclusion}, and human-factors engineering that makes symbolic provenance traces legible to clinicians without formal logic training.

\subsection{The Neuro-Symbolic Evaluation Deficit}
\label{subsec:ns_eval}

The clinical neuro-symbolic literature makes the most assertive safety claims among the paradigms reviewed here, yet its evaluation infrastructure is the least developed. Every quantitative result reviewed in the preceding sections is bounded by the coverage and correctness of the symbolic specification that produced it, and none of the reviewed studies systematically measures that bound. A system that asserts formal correctness guarantees whose scope it cannot quantify is not demonstrably safer than a well-calibrated neural agent, and its failure envelope is unknown.

Three evaluation dimensions remain unaddressed in the reviewed literature. First, none of the reviewed frameworks audits the symbolic layer's \textit{specification soundness}, meaning whether the encoded Prolog rules, OWL axioms, or KG relations faithfully formalize their biomedical source guidelines. Validating that takes clinical appraisal methods beyond standard NLP benchmarking. The reviewed architectures also lack metrics for \textit{veto precision and recall}, which would measure the fraction of hallucinated outputs that are intercepted against those that spread undetected. Song et al.~\cite{Song2025TrustEHRAgent} demonstrate reliability cliffs for unconstrained neural agents under strict reliability requirements (Section~\ref{subsubsec:calibration_cliff}), and in the same way a neuro-symbolic veto layer without independently measurable recall provides no auditable safety guarantee.

Second, the reviewed studies report no \textit{coverage boundary metrics}. Systems that silently pass out-of-domain queries to an unconstrained neural component have unmeasured blind spots, which hide performance degradation at the symbolic boundary and inflate published accuracy figures. Third, the reviewed frameworks do not measure \textit{temporal specification decay}, the rate at which clinical guideline updates supersede encoded rules. Architectures grounded in evolving medical protocols, such as the reinforcement learning alignment to ACR criteria of Tziakouri and Menolascina~\cite{Tziakouri2025GRPORL} and the clinical ontology of Ali et al.~\cite{OntologyCOVID2026}, are prone to going out of date, and neither offers a method to measure or reduce that decay during long-term deployment.

The evaluation infrastructure that would address these gaps, which we can call a NeuroSymBench, would invert the logic of MedPRMBench~\cite{MedPRMBench2026} (Section~\ref{subsec:eval_bottleneck}). MedPRMBench evaluates whether a trained model correctly identifies step-level errors. NeuroSymBench would evaluate whether the symbolic specification is clinically correct and current enough to justify the guarantees claimed for it. Its main artifact would be a versioned corpus of formally specified clinical knowledge structures, paired with expert correctness certificates, coverage maps, and guideline-update histories, against which specification staleness can be computed. It would fill, at the level of the symbolic layer, the same kind of gap as the outcome-measurement problem this review identifies at the level of the neural layer (Section~\ref{subsec:static_failure}). That infrastructure does not yet exist, and until it does, the safety advantages of neuro-symbolic grounding remain qualitatively compelling but quantitatively unaudited.

\section{Conclusion}
\label{sec:conclusion}

This review has traced an arc in the clinical AI literature. Each architectural response resolves an immediate bottleneck and exposes a deeper one. Static models hallucinate with confidence, and the metrics used to evaluate them often cannot detect it. Agentic architectures copied the iterative logic of clinical practice but introduced a more dangerous failure mode, the hallucination cascade, in which later steps inherit an early error as verified fact. Process Reward Models made reasoning steps visible by swapping sparse outcome rewards for dense step-level signals, yet their reliance on probabilistic neural verification points to the need for formal guarantees. Neuro-symbolic grounding is the most direct route to such guarantees in the reviewed literature. Formal logic engines, ontology reasoners, knowledge graphs, and theorem provers enforce explicit constraints over neural outputs without depending on parametric self-correction. Within their encoded domains, systems with symbolic veto layers offer guarantees conditional on their encoded knowledge and report large gains on the safety measures they use (Section~\ref{sec:neurosymbolic}). These results come mostly from single studies in simulated or small-sample, closed-world settings, and they are not direct cross-paradigm comparisons: experimental settings, task definitions, and hallucination operationalizations differ, so the synthesis here is structural and architectural, not meta-analytic.

The argument that emerges from this four-layer analysis is that clinical AI safety is to a large extent an architecture problem, and not only a matter of training, prompting, or calibration. The hallucination failures of Section~\ref{sec:core_problem} trace back largely to training objectives that reward fluency over fidelity and to evaluation pipelines that measure outcomes instead of processes. The cascade failures of Section~\ref{sec:agentic} trace back to pipeline architectures that treat every intermediate output as if it were verified clinical evidence. We interpret the calibration cliff of TrustEHRAgent, in which agents held to a 90\% reliability requirement have a very low correct-answer yield~\cite{Song2025TrustEHRAgent}, as a structural consequence of a measurable mismatch between the confidence an agent reports, the reward it was trained on, and how often it is actually correct, which we hypothesize tuning alone is unlikely to remove. Process supervision and neuro-symbolic grounding act as architectural corrections, and how well they work depends on how deeply they are built into the system's reasoning infrastructure, as opposed to applied as post-hoc filters.

Three open problems qualify these conclusions. The first is temporal adaptability. Every formal verification system is bounded by the currency of its symbolic specification, and because clinical guidelines change quickly, static knowledge representations carry silent, accumulating safety risks. GLEAN's Bayesian calibration~\cite{GLEAN2026} and Med-PRM's retrieval-augmented rewards~\cite{Yun2025MedPRM} (Section~\ref{subsubsec:discriminative_prm}) offer preliminary mitigations, but none of the reviewed architectures automates continuous adaptation. The second is generalization across clinical scope. The reviewed neuro-symbolic systems remain strictly domain-specific, and the literature offers neither a universally comprehensive biomedical ontology nor a principled orchestration theory for routing across domain-specific verifiers (Section~\ref{subsubsec:ns_tradeoffs}). The third is that process evaluation infrastructure lags system development. MedPRMBench~\cite{MedPRMBench2026} advances the field with a four-level severity grading system, but the broader literature's reliance on static, domain-bounded benchmarks leaves the true safety profiles of most neuro-symbolic architectures substantially undercharacterized, a gap Section~\ref{subsec:ns_eval} treats at length.

These open problems add to the review's own scope limitations. Purposive sampling over-represents architectures with clear mechanistic accounts, and preprint-sourced findings may not replicate under peer review. The two-batch search design (8 and 28 April 2026) introduces a minor temporal asymmetry, and papers published since may already address problems identified here. The integration architecture proposed in Section~\ref{sec:neurosymbolic} has had its components validated only in isolation, so its safety claims are architectural and not yet empirical. Several verification mechanisms discussed here were evaluated only outside the clinical domain (marked \dag\ in Tables~\ref{tab:paradigm_characteristics} and~\ref{tab:verification_regimes}), so their clinical benefit is inferred. Literature identification was supported by AI-assisted retrieval tools, and all retrieved studies were manually verified against indexed or publisher sources. Every quantitative finding and safety guarantee reported here holds only within the specific patient populations, evaluation settings, and encoded knowledge coverage of the primary studies cited.

These limitations also have regulatory and deployment implications. Current regulatory frameworks, including the FDA AI/ML SaMD Action Plan~\cite{FDA2021SaMD}, the EU AI Act 2024/1689~\cite{EUAIAct2024}, and the MHRA guidance~\cite{MHRA2023SaMD}, focus mainly on performance, risk management, and post-market monitoring. They are not purely outcome-based: for high-risk systems, the EU AI Act also requires risk management, technical documentation, automatic record-keeping, transparency towards deployers, and human oversight (Articles 9 to 14), alongside conformity assessment (Article 43). None of them, however, requires formal verification of the reasoning steps behind each output, or abstention when such verification fails. Responsible certification would require tracing every recommendation to its inferential pathway, verifying that pathway against formally specified knowledge and its premises against the source record, and confirming appropriate abstention on failure. The neuro-symbolic architectures reviewed here are not yet generalizable, standardized, or governed enough to support those capacities. Three practical risks follow directly. \textit{Automation bias}: a clinician who sees a ``verified'' label may give it more epistemic authority than the verifier's actual false-negative rate warrants, which turns a bounded safety margin into a trust amplifier for undetected errors. \textit{Liability assignment}: if a symbolic verifier emits a false negative that contributes to patient harm, how responsibility is shared among the neural model developer, the symbolic ontology curator, and the deploying institution is likely to be uncertain and to vary by jurisdiction. \textit{Latency and legibility}: the inference-time overhead of Prolog backward-chaining, OWL satisfiability testing, and multi-hop knowledge graph validation has not been evaluated against emergency or intensive-care time constraints. The provenance traces these methods produce are formally precise, but they are not legible at the clinical interface without translation into interruptible, human-readable annotations, and the fact that near-term use will be clinician-supervised rather than autonomous does not close that gap.

Viewed across the literature reviewed here, the field is moving from distributional fluency toward verifiable clinical logic, from sparse outcome rewards toward dense process supervision, and from opaque neural generation toward transparent, auditable neuro-symbolic reasoning. The systems at its frontier remain domain-constrained, computationally expensive, and insufficiently evaluated (Section~\ref{subsec:ns_eval}). Closing that evaluation gap, more than building a larger language model or a more sophisticated agent, will determine whether the guarantees documented in this review extend beyond the closed-world conditions in which they were demonstrated.

\section*{Ethics Statement}
This study did not involve experiments with human or animal subjects.

\section*{Acknowledgment}
The authors would like to thank Prince Sultan University and the University of Prince Mugrin for their support.

\section*{Funding}
This research did not receive any specific grant from funding agencies in the public, commercial, or not-for-profit sectors.

\section*{Declaration of competing interest}
The authors declare that they have no known competing financial interests or personal relationships that could have appeared to influence the work reported in this paper.

\section*{CRediT authorship contribution statement}
\textbf{Youssef ElNahas:} Conceptualization, Methodology, Investigation, Writing – original draft. \textbf{Soumaya Chaffar:} Conceptualization, Methodology, Investigation, Writing – editing. \textbf{Adel Ammar:} Conceptualization, Methodology, Investigation, Validation, Writing – editing. \textbf{Wadii Boulila:} Conceptualization, Methodology, Investigation, Validation, Writing – editing.

\section*{Data availability}
The list of included studies, with their paradigm assignments, the version cited and, where different, the version screened, is provided in Supplementary Table~S1. No new primary data were generated in this study.

\section*{Declaration of generative AI and AI-assisted technologies in the manuscript preparation process}
During the preparation of this work the authors used Perplexity AI, Claude (Anthropic), DeepSeek, Grok (xAI), and Kimi to support structured literature identification, as described in Section~\ref{sec:methods}. The authors also used Claude (Anthropic) and ChatGPT to check the manuscript for consistency with its cited sources and to suggest revisions to wording and structure. After using these tools, the authors manually verified all retrieved records against indexed sources, reviewed and edited the content as needed, and assume full responsibility for the content of the present article.

\appendix

\section{List and definitions of abbreviations/acronyms used in this review}
\begin{table}[H]
\begin{tabular}{l l}
\hline
\textbf{Abbreviations/Acronyms} & \textbf{Definition} \\ \hline
AI    & Artificial Intelligence \\
ALP   & Abductive Logic Programming \\
CLP   & Constraint Logic Programming \\
DL    & Deep Learning; Description Logic (in OWL ontology contexts) \\
EHR   & Electronic Health Record \\
FHIR  & Fast Healthcare Interoperability Resources \\
GRPO  & Group-Relative Policy Optimization \\
KG    & Knowledge Graph \\
LLM   & Large Language Model \\
MAD   & Multi-Agent Debate \\
MCQ   & Multiple Choice Question \\
MCTS  & Monte Carlo Tree Search \\
ORM   & Outcome Reward Model \\
OWL   & Web Ontology Language \\
PRA   & Process Reward Agent \\
PRM   & Process Reward Model \\
RLHF  & Reinforcement Learning from Human Feedback \\
SaMD  & Software as a Medical Device \\
SPARQL & SPARQL Protocol and RDF Query Language \\
VPRM  & Verifiable Process Reward Model \\
\hline
\end{tabular}
\end{table}

 \bibliographystyle{elsarticle-num} 
 \bibliography{cas-refs}

\end{document}

\noindent{\large\textbf{Supplementary Material}}\\[2pt]
\noindent Verifiable Logic in Medical LLMs: A Critical Review of Agentic Clinical Reasoning Optimization and Neuro-Symbolic Grounding\\[8pt]
\noindent\textbf{Supplementary Table S1.} Primary studies included in the review ($n=98$), grouped by research paradigm. \emph{Ref.}\ is the reference number in the main manuscript. \emph{Type} is the source type at retrieval (P = peer-reviewed; PP = preprint; CF = conceptual or speculative framework). Where a preprint was published after the search closed (28 April 2026), the version cited is the published one and the version screened is given in a note. Records assessed and included per paradigm: RQ1, 39 and 29; RQ2, 38 and 23; RQ3, 30 and 24; RQ4, 31 and 22 (138 records assessed in total after two records logged under two research questions were merged).
\vspace{4pt}
\begin{longtable}{>{\raggedleft\arraybackslash}p{0.5cm} >{\raggedleft\arraybackslash}p{0.7cm} >{\raggedright\arraybackslash}p{3.3cm} >{\raggedright\arraybackslash}p{9.6cm} >{\raggedright\arraybackslash}p{4.6cm} >{\raggedright\arraybackslash}p{4.4cm} >{\centering\arraybackslash}p{0.8cm}}
\toprule
\textbf{No.} & \textbf{Ref.} & \textbf{Study} & \textbf{Title} & \textbf{Version cited} & \textbf{Identifier} & \textbf{Type}\\
\midrule\endfirsthead
\toprule
\textbf{No.} & \textbf{Ref.} & \textbf{Study} & \textbf{Title} & \textbf{Version cited} & \textbf{Identifier} & \textbf{Type}\\
\midrule\endhead
\bottomrule\endfoot
\multicolumn{7}{l}{\textbf{RQ1: Hallucination taxonomy and benchmark failure}}\\[1pt]
1 & 27 & Kim et al. (2025) & Limitations of large language models in clinical problem-solving arising from inflexible reasoning & Scientific Reports & \url{10.1038/s41598-025-22940-0} & P\\
2 & 28 & Zi et al. (2026) & Shattering the Shortcut: A Topology-Regularized Benchmark for Multi-hop Medical Reasoning in LLMs & arXiv preprint & \url{10.48550/arXiv.2603.12458} & PP\\
3 & 29 & Pandit et al. (2025) & MedHallu: A Comprehensive Benchmark for Detecting Medical Hallucinations in Large Language Models & arXiv preprint & \url{10.48550/arXiv.2502.14302} & PP\\
4 & 30 & Wu et al. (2025) & MedCaseReasoning: Evaluating and Learning Diagnostic Reasoning from Clinical Case Reports & arXiv preprint & \url{arXiv:2505.11733} & PP\\
5 & 39 & Omar et al. (2025) & Sociodemographic biases in medical decision making by large language models & Nature Medicine & \url{10.1038/s41591-025-03626-6} & P\\
6 & 40 & Kim et al. (2025) & Medical Hallucinations in Foundation Models and Their Impact on Healthcare & arXiv preprint & \url{10.48550/arXiv.2503.05777} & PP\\
7 & 41 & Asgari et al. (2025) & A framework to assess clinical safety and hallucination rates of LLMs for medical text summarisation & npj Digital Medicine & \url{10.1038/s41746-025-01670-7} & P\\
8 & 42 & Williams et al. (2025) & Evaluating large language models for drafting emergency department encounter summaries & PLOS Digital Health & \url{10.1371/journal.pdig.0000899} & P\\
9 & 43 & Normand et al. (2025) & A Real-World Evaluation of LLM Medication Safety Reviews in NHS Primary Care & arXiv preprint & \url{10.48550/arXiv.2512.21127} & PP\\
10 & 44 & Omar et al. (2025) & Multi-model assurance analysis showing large language models are highly vulnerable to adversarial hallucination attacks during clinical decision support & Communications Medicine & \url{10.1038/s43856-025-01021-3} & P\\
11 & 45 & Li et al. (2026) & Beyond Idealized Patients: Evaluating LLMs under Challenging Patient Behaviors in Medical Consultations & arXiv preprint & \url{10.48550/arXiv.2603.29373} & PP\\
12 & 46 & Griot et al. (2025) & Large Language Models lack essential metacognition for reliable medical reasoning & Nature Communications & \url{10.1038/s41467-024-55628-6} & P\\
13 & 47 & Omar et al. (2025) & Benchmarking the Confidence of Large Language Models in Answering Clinical Questions: Cross-Sectional Evaluation Study & JMIR Medical Informatics & \url{10.2196/66917} & P\\
14 & 48 & Savage et al. (2025) & Large language model uncertainty proxies: discrimination and calibration for medical diagnosis and treatment & Journal of the American Medical Informatics Association & \url{10.1093/jamia/ocae254} & P\\
15 & 49 & Au Yeung et al. (2025) & The Psychogenic Machine: Simulating AI Psychosis, Delusion Reinforcement and Harm Enablement in Large Language Models & arXiv preprint & \url{10.48550/arXiv.2509.10970} & PP\\
16 & 50 & Chen et al. (2025) & CARES: Comprehensive Evaluation of Safety and Adversarial Robustness in Medical LLMs & arXiv preprint & \url{10.48550/arXiv.2505.11413} & PP\\
17 & 51 & Das et al. (2025) & Trustworthy Medical Imaging with Large Language Models: A Study of Hallucinations Across Modalities & arXiv preprint & \url{10.48550/arXiv.2508.07031} & PP\\
18 & 52 & Gu et al. (2024) & MedVH: Towards Systematic Evaluation of Hallucination for Large Vision Language Models in the Medical Context & arXiv preprint & \url{10.48550/arXiv.2407.02730} & PP\\
19 & 53 & Chang et al. (2025) & MedHEval: Benchmarking Hallucinations and Mitigation Strategies in Medical Large Vision-Language Models & arXiv preprint & \url{10.48550/arXiv.2503.02157} & PP\\
20 & 54 & Chen et al. (2024) & Detecting and Evaluating Medical Hallucinations in Large Vision Language Models & arXiv preprint & \url{10.48550/arXiv.2406.10185} & PP\\
21 & 55 & Zuo et al. (2025) & MedHallBench: A New Benchmark for Assessing Hallucination in Medical Large Language Models & arXiv preprint & \url{arXiv:2412.18947} & PP\\
22 & 56 & Atwany et al. (2025) & Lost in Transcription, Found in Distribution Shift: Demystifying Hallucination in Speech Foundation Models & arXiv preprint & \url{10.48550/arXiv.2502.12414} & PP\\
23 & 59 & Chen et al. (2025) & Beyond the Leaderboard: Rethinking Medical Benchmarks for Large Language Models & arXiv preprint & \url{arXiv:2508.04325} & PP\\
24 & 60 & Mruthyunjaya et al. (2025) & Right Diagnoses But Wrong Reasoning: Current Large-Language Model-Based Agentic Frameworks Have Flawed Clinical Reasoning Despite High Diagnostic Accuracy & SSRN Electronic Journal & \url{10.2139/ssrn.5339074} & PP\\
25 & 61 & Roustan et al. (2025) & The Clinicians' Guide to Large Language Models: A General Perspective With a Focus on Hallucinations & Interactive Journal of Medical Research & \url{10.2196/59823} & P\\
26 & 62 & Stamatis et al. (2026) & Beyond Simulations: What 20,000 Real Conversations Reveal About Mental Health AI Safety & arXiv preprint & \url{arXiv:2601.17003} & PP\\
27 & 63 & Pan et al. (2025) & Addressing Benchmarking Gaps in Large Language Models for Health and Medicine with Dynamic Red-Teaming & arXiv preprint & \url{10.48550/arXiv.2508.00923} & PP\\
28 & 64 & Agarwal et al. (2024) & MedHalu: Hallucinations in Responses to Healthcare Queries by Large Language Models & arXiv preprint & \url{10.48550/arXiv.2409.19492} & PP\\
29 & 65 & Pawlik et al. (2026) & Reducing Hallucinations in Medical AI Through Citation Enforced Prompting in RAG Systems & Applied Sciences & \url{10.3390/app16063013} & P\\
\multicolumn{7}{l}{\textbf{RQ2: Agentic architectures and cascade failure modes}}\\[1pt]
30 & 31 & Shi et al. (2024) & EHRAgent: Code Empowers Large Language Models for Few-shot Complex Tabular Reasoning on Electronic Health Records & Proceedings of the 2024 Conference on Empirical Methods in Natural Language Processing & \url{10.18653/v1/2024.emnlp-main.1245} & P\\
31 & 32 & Jiang et al. (2025) & MedAgentBench: A Realistic Virtual EHR Environment to Benchmark Medical LLM Agents & NEJM AI & \url{arXiv:2501.14654} & P\\
32 & 33 & Zhu et al. (2025) & Auditing medical multi-agent AI reveals risks of false consensus & arXiv preprint & \url{10.48550/arXiv.2510.10185} & PP\\
33 & 34 & Song et al. (2025) & Trustworthy Agents for Electronic Health Records through Confidence Estimation & arXiv preprint & \url{arXiv:2508.19096} & PP\\
34 & 66 & Liao et al. (2026) & AgentEHR: Advancing Autonomous Clinical Decision-Making via Retrospective Summarization & arXiv preprint & \url{10.48550/arXiv.2601.13918} & PP\\
35 & 70 & Tang et al. (2024) & MedAgents: Large Language Models as Collaborators for Zero-shot Medical Reasoning & Findings of the Association for Computational Linguistics: ACL & \url{arXiv:2311.10537} & P\\
36 & 71 & Wang et al. (2025) & ColaCare: Enhancing Electronic Health Record Modeling through Large Language Model-Driven Multi-Agent Collaboration & Proceedings of the ACM on Web Conference 2025 (WWW '25) & \url{10.1145/3696410.3714877} & P\\
37 & 72 & Li et al. (2024) & MMedAgent: Learning to Use Medical Tools with Multi-modal Agent & arXiv preprint & \url{10.48550/arXiv.2407.02483} & PP\\
38 & 73 & Li et al. (2026)\textsuperscript{a} & Empowering clinical trial design with agentic intelligence and real-world data & Nature Communications & \url{10.1038/s41467-026-74501-2} & PP\\
39 & 74 & Schmidgall et al. (2024) & AgentClinic: A Multimodal Agent Benchmark to Evaluate AI in Simulated Clinical Environments & arXiv preprint & \url{arXiv:2405.07960} & PP\\
40 & 75 & Liu et al. (2026) & Benchmarking large language model-based agent systems for clinical decision tasks & npj Digital Medicine & \url{10.1038/s41746-026-02443-6} & P\\
41 & 76 & Basu (2026) & Tool Receipts, Not Zero-Knowledge Proofs: Practical Hallucination Detection for AI Agents & arXiv preprint & \url{10.48550/arXiv.2603.10060} & PP\\
42 & 77 & Masayoshi et al. (2025) & EHR-MCP: Real-world Evaluation of Clinical Information Retrieval by Large Language Models via Model Context Protocol & arXiv preprint & \url{10.48550/arXiv.2509.15957} & PP\\
43 & 78 & Bi et al. (2025) & MAGI: Multi-Agent Guided Interview for Psychiatric Assessment & arXiv preprint & \url{10.48550/arXiv.2504.18260} & PP\\
44 & 79 & Li et al. (2025) & MACD: Multi-Agent Clinical Diagnosis with Self-Learned Knowledge for LLM & arXiv preprint & \url{10.48550/arXiv.2509.20067} & PP\\
45 & 80 & Wang et al. (2026)\textsuperscript{b} & MediHive: A Decentralized Agent Collective for Medical Reasoning & Journal of Healthcare Informatics Research & \url{10.1007/s41666-026-00239-7} & PP\\
46 & 81 & Jeong et al. (2026) & Tool-MAD: A Multi-Agent Debate Framework for Fact Verification with Diverse Tool Augmentation and Adaptive Retrieval & arXiv preprint & \url{10.48550/arXiv.2601.04742} & PP\\
47 & 83 & Cao et al. (2026) & EHR-RAG: Bridging Long-Horizon Structured Electronic Health Records and Large Language Models via Enhanced Retrieval-Augmented Generation & arXiv preprint & \url{10.48550/arXiv.2601.21340} & PP\\
48 & 84 & Mu et al. (2026) & EHRWorld: A Patient-Centric Medical World Model for Long-Horizon Clinical Trajectories & arXiv preprint & \url{10.48550/arXiv.2602.03569} & PP\\
49 & 85 & Ge et al. (2026) & ClinicalAgents: Multi-Agent Orchestration for Clinical Decision Making with Dual-Memory & arXiv preprint & \url{10.48550/arXiv.2603.26182} & PP\\
50 & 86 & Xu et al. (2025) & AutoCBT: An Autonomous Multi-agent Framework for Cognitive Behavioral Therapy in Psychological Counseling & arXiv preprint & \url{10.48550/arXiv.2501.09426} & PP\\
51 & 87 & Ozgun et al. (2025) & Trustworthy AI Psychotherapy: Multi-Agent LLM Workflow for Counseling and Explainable Mental Disorder Diagnosis & arXiv preprint & \url{10.48550/arXiv.2508.11398} & PP\\
52 & 88 & Badawi et al. (2026) & Assessing the Quality of Mental Health Support in LLM Responses through Multi-Attribute Human Evaluation & arXiv preprint & \url{10.48550/arXiv.2601.18630} & PP\\
\multicolumn{7}{l}{\textbf{RQ3: Process reward model verification}}\\[1pt]
53 & 16 & Yun et al. (2025) & Med-PRM: Medical Reasoning Models with Stepwise, Guideline-verified Process Rewards & Proceedings of the 2025 Conference on Empirical Methods in Natural Language Processing & \url{arXiv:2506.11474} & P\\
54 & 17 & Sohn et al. (2026) & Process Reward Agents for Steering Knowledge-Intensive Reasoning & arXiv preprint & \url{10.48550/arXiv.2604.09482} & PP\\
55 & 19 & Wang et al. (2025) & Process-Supervised Reward Models for Verifying Clinical Note Generation: A Scalable Approach Guided by Domain Expertise & Proceedings of the 2025 Conference on Empirical Methods in Natural Language Processing & \url{10.18653/v1/2025.emnlp-main.967} & P\\
56 & 20 & Thomas et al. (2025) & Process Reward Models for Sentence-Level Verification of LVLM Radiology Reports & arXiv preprint & \url{10.48550/arXiv.2510.23217} & PP\\
57 & 21 & Wu et al. (2025) & MedReason: Eliciting Factual Medical Reasoning Steps in LLMs via Knowledge Graphs & arXiv preprint & \url{10.48550/arXiv.2504.00993} & PP\\
58 & 22 & Mu et al. (2025) & MedCEG: Reinforcing Verifiable Medical Reasoning with Critical Evidence Graph & arXiv preprint & \url{10.48550/arXiv.2512.13510} & PP\\
59 & 23 & Rui et al. (2025) & Improving Medical Reasoning with Curriculum-Aware Reinforcement Learning & arXiv preprint & \url{10.48550/arXiv.2505.19213} & PP\\
60 & 35 & Pronesti et al. (2026) & Beyond Outcome Verification: Verifiable Process Reward Models for Structured Reasoning & arXiv preprint & \url{10.48550/arXiv.2601.17223} & PP\\
61 & 36 & Jiang et al. (2026) & MedS\textsuperscript{3}: Towards Medical Slow Thinking with Self-Evolved Soft Dual-sided Process Supervision & Proceedings of the AAAI Conference on Artificial Intelligence & \url{10.1609/aaai.v40i37.40395} & P\\
62 & 89 & Pan et al. (2025) & MedVLM-R1: Incentivizing Medical Reasoning Capability of Vision-Language Models (VLMs) via Reinforcement Learning & arXiv preprint & \url{10.48550/arXiv.2502.19634} & PP\\
63 & 90 & Zhong et al. (2026) & P2S: Probabilistic Process Supervision for General-Domain Reasoning Question Answering & arXiv preprint & \url{10.48550/arXiv.2601.20649} & PP\\
64 & 91 & Liu et al. (2025) & Fleming-R1: Toward Expert-Level Medical Reasoning via Reinforcement Learning & arXiv preprint & \url{10.48550/arXiv.2509.15279} & PP\\
65 & 92 & Chen et al. (2024) & HuatuoGPT-o1, Towards Medical Complex Reasoning with LLMs & arXiv preprint & \url{10.48550/arXiv.2412.18925} & PP\\
66 & 94 & Xiao et al. (2025) & MiraMind: Benchmarking Reliable Mental Health Reasoning beyond Answer Accuracy & arXiv preprint & \url{10.48550/arXiv.2512.09636} & PP\\
67 & 95 & Dai et al. (2025) & Psyche-R1: Towards Reliable Psychological LLMs through Unified Empathy, Expertise, and Reasoning & arXiv preprint & \url{10.48550/arXiv.2508.10848} & PP\\
68 & 96 & Feng et al. (2026) & From Stimuli to Minds: Enhancing Psychological Reasoning in LLMs via Bilateral Reinforcement Learning & Proceedings of the AAAI Conference on Artificial Intelligence & \url{10.1609/aaai.v40i1.36988} & P\\
69 & 97 & Tarek et al. (2025) & Reward Hacking Mitigation using Verifiable Composite Rewards & arXiv preprint & \url{10.48550/arXiv.2509.15557} & PP\\
70 & 98 & Zhang et al. (2025) & Med-U1: Incentivizing Unified Medical Reasoning in LLMs via Large-scale Reinforcement Learning & arXiv preprint & \url{10.48550/arXiv.2506.12307} & PP\\
71 & 99 & Zheng et al. (2026) & MedPRMBench: A Fine-grained Benchmark for Process Reward Models in Medical Reasoning & arXiv preprint & \url{10.48550/arXiv.2604.17282} & PP\\
72 & 100 & Ding et al. (2025) & Med-RewardBench: Benchmarking Reward Models and Judges for Medical Multimodal Large Language Models & arXiv preprint & \url{10.48550/arXiv.2508.21430} & PP\\
73 & 101 & Zhou et al. (2026) & Automating expert-level medical reasoning evaluation of large language models & npj Digital Medicine & \url{10.1038/s41746-025-02208-7} & P\\
74 & 102 & Gupta et al. (2025) & LLMs in Disease Diagnosis: A Comparative Study of DeepSeek-R1 and O3 Mini Across Chronic Health Conditions & arXiv preprint & \url{10.48550/arXiv.2503.10486} & PP\\
75 & 103 & Tziakouri et al. (2025) & Reinforcement Learning for Clinical Reasoning: Aligning LLMs with ACR Imaging Appropriateness Criteria & arXiv preprint & \url{10.48550/arXiv.2510.05194} & PP\\
76 & 104 & Zhang et al. (2026) & Guideline-Grounded Evidence Accumulation for High-Stakes Agent Verification & arXiv preprint & \url{10.48550/arXiv.2603.02798} & PP\\
\multicolumn{7}{l}{\textbf{RQ4: Neuro-symbolic grounding}}\\[1pt]
77 & 13 & Tan et al. (2026) & Prolog-Driven Rule-Based Diagnostics with Large Language Models for Precise Clinical Decision Support & Medical Image Computing and Computer Assisted Intervention -- MICCAI 2025 & \url{10.1007/978-3-032-05127-1_40} & P\\
78 & 14 & Quan et al. (2024) & Verification and Refinement of Natural Language Explanations through LLM-Symbolic Theorem Proving & Proceedings of the 2024 Conference on Empirical Methods in Natural Language Processing & \url{10.18653/v1/2024.emnlp-main.172} & P\\
79 & 15 & Magana Vsevolodovna et al. (2025) & Enhancing Large Language Models through Neuro-Symbolic Integration and Ontological Reasoning & arXiv preprint & \url{10.48550/arXiv.2504.07640} & PP\\
80 & 18 & Ali et al. (2026) & Ontology-grounded knowledge graphs for mitigating hallucinations in large language models for clinical question answering & Journal of biomedical informatics & \url{10.1016/j.jbi.2026.104993} & P\\
81 & 37 & Sansford et al. (2024) & GraphEval: A Knowledge-Graph Based LLM Hallucination Evaluation Framework & Proceedings of the KiL Workshop at KDD 2024 & \url{arXiv:2407.10793} & P\\
82 & 108 & Kim et al. (2025) & Large Language Models for Interpretable Mental Health Diagnosis & arXiv preprint & \url{arXiv:2501.07653} & PP\\
83 & 109 & Arriola-Montenegro et al. (2025) & A deterministic large language model (LLM) framework for safe, protocol-adherent clinical decision support: application in hemodialysis anemia management (AnemiaCare HDs) & Frontiers in Artificial Intelligence & \url{10.3389/frai.2025.1728320} & P\\
84 & 112 & Pusch et al. (2025) & Combining LLMs and Knowledge Graphs to Reduce Hallucinations in Biomedical Question Answering & BioMedInformatics & \url{10.3390/biomedinformatics5040070} & P\\
85 & 113 & Wang et al. (2025) & MedKGI: Iterative Differential Diagnosis with Medical Knowledge Graphs and Information-Guided Inquiring & arXiv preprint & \url{10.48550/arXiv.2512.24181} & PP\\
86 & 114 & Cheng et al. (2026)\textsuperscript{c} & LogosKG: Hardware-Optimized Scalable and Interpretable Knowledge Graph Retrieval & Proceedings of the 64th Annual Meeting of the Association for Computational Linguistics (Volume 1: Long Papers) & \url{https://aclanthology.org/2026.acl-long.1966/} & PP\\
87 & 115 & Wu et al. (2025) & KG-TRACES: Enhancing Large Language Models with Knowledge Graph-constrained Trajectory Reasoning and Attribution Supervision & arXiv preprint & \url{10.48550/arXiv.2506.00783} & PP\\
88 & 116 & Hu et al. (2025) & A self-correcting Agentic Graph RAG for clinical decision support in hepatology & Frontiers in Medicine & \url{10.3389/fmed.2025.1716327} & P\\
89 & 117 & Wang et al. (2024) & RuleAlign: Making Large Language Models Better Physicians with Diagnostic Rule Alignment & arXiv preprint & \url{10.48550/arXiv.2408.12579} & PP\\
90 & 118 & An et al. (2025) & A Neurosymbolic Approach to Natural Language Formalization and Verification & arXiv preprint & \url{10.48550/arXiv.2511.09008} & PP\\
91 & 119 & Galitsky et al. (2026) & Neuro-Symbolic Verification for Preventing LLM Hallucinations in Process Control & Processes & \url{10.3390/pr14020322} & P\\
92 & 120 & Zhang et al. (2025) & Haibu Mathematical-Medical Intelligent Agent: Enhancing Large Language Model Reliability in Medical Tasks via Verifiable Reasoning Chains & arXiv preprint & \url{10.48550/arXiv.2510.07748} & PP\\
93 & 121 & Wang et al. (2026) & Explainable neuro-symbolic artificial intelligence for automated interpretation of corneal topography and early keratoconus detection & Frontiers in Artificial Intelligence & \url{10.3389/frai.2026.1713747} & P\\
94 & 122 & Hong et al. (2026) & Adaptive diagnostic reasoning framework for pathology with multimodal large language models & Communications Medicine & \url{10.1038/s43856-026-01491-z} & P\\
95 & 123 & Lopukhova et al. (2025) & An Interpretable Clinical Decision Support System Aims to Stage Age-Related Macular Degeneration Using Deep Learning and Imaging Biomarkers & Applied Sciences & \url{10.3390/app151810197} & P\\
96 & 124 & Kiruluta (2025) & A Novel Architecture for Symbolic Reasoning with Decision Trees and LLM Agents & arXiv preprint & \url{arXiv:2508.05311} & PP\\
97 & 125 & Bhattacharya et al. (2025) & Neurosymbolic AI for Safe and Trustworthy High-Stakes Applications & Preprints & \url{10.20944/preprints202511.1342.v1} & CF\\
98 & 126 & Kolli (2025) & Hybrid Neuro-Symbolic Models for Ethical AI in Risk-Sensitive Domains & arXiv preprint & \url{10.48550/arXiv.2511.17644} & CF\\
\end{longtable}
\noindent\textit{Version screened:}
\textsuperscript{a}medRxiv preprint, doi:10.1101/2025.04.17.25326033 (2025); published 2026. 
\textsuperscript{b}arXiv:2603.27150 (March 2026); journal version published online 27 May 2026. 
\textsuperscript{c}medRxiv preprint, doi:10.64898/2026.01.12.26343957 (January 2026), titled "Scaling Biomedical Knowledge Graph Retrieval for Interpretable Reasoning: Applications to Clinical Diagnosis Prediction"; published at ACL 2026.